\documentclass[11pt]{article}
\usepackage{graphicx,psfrag,epsf}
\usepackage{enumerate}
\usepackage{natbib}
\usepackage{url} 
\graphicspath{{C:/Users/whsigris/Dropbox/HSLU/Projects/KTI_FraudDetection_CreditRisk/boosting_comparison/article/plots_tables/}}
\usepackage{amssymb, amsmath, amsthm,graphicx,bbm,float,bm}
\usepackage[linesnumbered,ruled]{algorithm2e}
\usepackage{color}
\usepackage[toc,page]{appendix}

\DeclareMathOperator*{\argmin}{argmin}
\newcommand{\vect}[1]{\bm{#1}}

\newcommand{\blind}{0}

\begin{document}

\newif\ifESWA
\ESWAfalse

\global\let\clearpagegood\clearpage
\global\let\clearpage\relax

\def\spacingset#1{\renewcommand{\baselinestretch}%
{#1}\small\normalsize} \spacingset{1}

\if0\blind
{
  \title{\bf Gradient and Newton Boosting for Classification and Regression}
  
  \author{Fabio Sigrist\thanks{Email: fabio.sigrist@hslu.ch. Address: Lucerne University of Applied Sciences and Arts, Suurstoffi 1, 6343 Rotkreuz, Switzerland. Phone: +41 41 757 67 61.}\\
   Lucerne University of Applied Sciences and Arts}
  \maketitle
}
\fi

\bigskip

\ifESWA
\newpage
\fi

\begin{abstract}
Boosting algorithms are frequently used in applied data science and in research. To date, the distinction between boosting with either gradient descent or second-order Newton updates is often not made in both applied and methodological research, and it is thus implicitly assumed that the difference is irrelevant. The goal of this article is to clarify this situation. In particular, we present gradient and Newton boosting, as well as a hybrid variant of the two, in a unified framework. We compare these boosting algorithms with trees as base learners using various datasets and loss functions. Our experiments show that Newton boosting outperforms gradient and hybrid gradient-Newton boosting in terms of predictive accuracy on the majority of datasets. We also present evidence that the reason for this is not faster convergence of Newton boosting. In addition, we introduce a novel tuning parameter for tree-based Newton boosting which is interpretable and important for predictive accuracy.
	
\end{abstract}

\noindent%

\spacingset{1.45} 

\global\let\clearpage\clearpagegood

\section{Introduction}\label{intro}

Boosting \citep{freund1996experiments, friedman2000additive, friedman2001greedy} refers to a type of supervised learning algorithms that enjoy high popularity in applied data science and research, among other things, due to their high predictive accuracy \citep{chen2016xgboost}. This is reflected in statements such as ``[i]n general `boosted decision trees' is regarded as the most effective off-the-shelf nonlinear learning method for a wide range of application problems"  \citep{johnson2013learning}. Boosting iteratively adds so-called base learners to an ensemble of learners. Broadly speaking, there exist three different versions for selecting a base learner in every boosting iteration: functional gradient descent, a functional version of Newton's method, and a combination of the two. We refer to these three different versions of boosting as \textit{gradient boosting}, \textit{Newton boosting}, and \textit{hybrid gradient-Newton boosting}; see Section \ref{boostpres} for more information.

\ifESWA
In expert and intelligent systems, tree-boosting is a technique that is frequently used. Recent applications of boosting in expert systems include bankruptcy prediction and credit scoring \citep{wang2014improved, xia2017boosted, djeundje2020enhancing, moscatelli2020corporate}, network intrusion detection \citep{zhou2020m}, epileptic seizure diagnosis \citep{al2020adaptive}, early stage disease symptom detection \citep{ahamad2020machine}, cancer prognosis \citep{lu2019dynamic}, and face re-identification \citep{soleymani2018progressive}.
\fi 

In both methodological and applied research, the distinction between gradient and Newton boosting is often not made and/or it is not declared which version of boosting is used \citep[e.g.][]{ahamad2020machine, djeundje2020enhancing, moscatelli2020corporate}. It is thus implicitly assumed that the difference is not important. For instance, the two recent popular boosting libraries  \texttt{LightGBM} and \texttt{TF Boosted Trees} do not distinguish in their companion articles \citep{ke2017lightgbm,ponomareva2017tf} between gradient and Newton boosting, and it is unclear to the reader which version is used. Similarly, \citet{Prokhorenkova_CatBoost} briefly mention in their article on \texttt{CatBoost} that the minimization for finding a boosting update can be done using the Newton method or with a gradient step, and then continue to write that ``[b]oth methods are kinds of functional gradient descent''. However, Newton's method is different from gradient descent. Further, \citet{buhlmann2007boosting} state that for gradient boosting ``an additional line search ... seems unnecessary for achieving a good estimator." For trees as base learners, the additional line search is often done for each leaf separately by using a Newton step \citep{friedman2001greedy}. I.e., this corresponds to what we denote as hybrid gradient-Newton boosting which is different from plain gradient boosting also in terms of predictive accuracy. Besides, particular software implementations of boosting such as \texttt{XGBoost} \citep{chen2016xgboost} are sometimes presented as if they were separate boosting algorithms \citep[e.g.][]{xia2017boosted,ahamad2020machine, djeundje2020enhancing} when, in fact, they implement a particular version of boosting. 

The novel contributions of this article are the following ones. First, we show how gradient, Newton, as well as hybrid gradient-Newton boosting can be derived in a unified framework. Further, we systematically compare gradient, Newton, and hybrid gradient-Newton boosting on a large set of both real-world and simulated classification and regression datasets. In our experiments, using trees as base learners, we find that Newton boosting achieves lower test errors than both gradient boosting and hybrid gradient-Newton boosting, and hybrid gradient-Newton boosting often has higher predictive accuracy than gradient boosting. Interestingly, we find that Newton boosting results in both lower in-sample training losses, which are essentially zero for most classification datasets, and lower out-of-sample test errors for most datasets. We also present evidence that the higher predictive accuracy is not due to a faster convergence speed of Newton boosting. In addition, we introduce a novel tuning parameter for Newton boosting with trees as base learners. We argue that this minimum \textit{equivalent sample size per leaf} parameter is a natural and interpretable tuning parameter which is important for predictive accuracy. In particular, we present evidence that the unnormalized version of this tuning parameter currently adopted in popular software implementations such as \texttt{XGBoost} is difficult to tune and thus likely results in lower predictive accuracy.

\subsection{Related work}
The first boosting algorithms for classification, including the well known AdaBoost algorithm, were introduced by \citet{schapire1990strength}, \citet{freund1995desicion}, and \citet{freund1996experiments}. Later, several authors \citep{breiman1998arcing,breiman1999prediction,friedman2000additive,mason2000boosting,friedman2001greedy} presented the statistical view of boosting as a stagewise optimization approach. See \citet{schapire2003boosting}, \citet{buhlmann2007boosting}, \citet{schapire2012boosting}, \citet{mayr2014evolution}, and \cite{mayr2014extending} for reviews on boosting algorithms in both the machine learning and statistical literature.

To the best of our knowledge, a systematic comparison concerning the predictive accuracy of gradient, Newton, and hybrid gradient-Newton boosting for various choices of loss functions, including regression and classification losses, has not been done so far. The $L_K$\_TreeBoost algorithm \citep{friedman2001greedy} is compared in \citet{friedman2001greedy} with $K$-class LogitBoost \citep{friedman2000additive} for classification with five classes in a simulation study for one type of random functions. In our terminology, $L_K$\_TreeBoost is a version of hybrid gradient-Newton boosting, and $K$-class LogitBoost corresponds to Newton boosting for the Bernoulli likelihood. \citet{friedman2001greedy} finds that the algorithms perform ``nearly the same'' with ``LogitBoost perhaps having a slight advantage''. In addition, it is mentioned that ``it is likely that when the shrinkage parameter is carefully tuned for each of the three methods [$L_K$\_TreeBoost, $K$-class LogitBoost, AdaBoost], there would be little performance differential between them.'' Our empirical evidence is not in line with this statement. \citet{saberian2011taylorboost} also briefly compare variants of boosting with gradient and second-order updates using three different binary classification datasets and Haar wavelets as base learners. However, their boosting approach is different from the one usually adopted in practice and research in the sense that they assume normed based learners, find base learners as maxima of inner products of gradients and base learners, and then have to perform an additional line search to find the step size. Further, tuning parameters such as the learning rate and the number of boosting iterations are not chosen using cross-validation, and only 25 boosting iterations are performed. Nonetheless, they come to the same conclusion as we do, i.e., they find that their version of second-order boosting performs better than gradient boosting. The closest to our empirical analysis are \citet{Li2010RobustLA} and \citet{zheng2012functional}. \citet{Li2010RobustLA} compares Newton boosting (``logitboost'') with hybrid gradient-Newton boosting (``mart'') for several multi-class classification datasets and also finds that Newton results in lower test errors than hybrid gradient-Newton boosting. Further, \citet{zheng2012functional} compare gradient and Newton boosting when using the probit link function in a logistic regression model and find that Newton boosting results in lower error rates than gradient boosting for several classification applications including face detection, cancer classification, and handwritten digit recognition. However, both \citet{Li2010RobustLA} and \citet{zheng2012functional} consider only specific classification tasks, tuning parameters are not chosen using validation data in their experiments, and it is not investigated whether the observed differences are statistically significant. Finally, \citet{sun2014convergence} compare Newton and gradient boosting for binary classification using the logistic loss. Their focus is on the convergence rate and their empirical comparison only considers training errors, though.

\section{The statistical view of boosting: three approaches for stagewise optimization}\label{boostpres}
In this section, we present the statistical view of boosting as finding the minimizer of a risk functional in a function space using a stagewise, or greedy, optimization approach. We distinguish between gradient and Newton boosting as wells as a hybrid version of the two and show how these can be presented in a unified framework. Note that these boosting algorithms have been proposed in prior research \citep{breiman1998arcing,breiman1999prediction,friedman2000additive,mason2000boosting, friedman2001greedy,saberian2011taylorboost}, but, to the best of our knowledge, the presentation below in a unified framework and the extension to the multivariate case is novel.

\subsection{Population versions}
We assume that there is a response variable $Y\in \mathbb{R}$ and a vector of $p$ predictor variables $X\in \mathbb{R}^p$.\footnote{For the sake of simplicity, we focus on univariate $Y\in \mathbb{R}$. The extension to the case of a multivariate response variable $Y$ is straightforward. See also Section \ref{mutlivar} where we present multivariate versions of boosting.} Our goal is to predict the response variable using the predictor variables, where predictions can be both deterministic or probabilistic. We assume that $(Y,X)$ are random variables on $\mathbb{R}\times\mathbb{R}^p$, and both the distribution of $X$ and the conditional distribution $Y|X$ are absolutely continuous with respect to either the Lebesgue measure, a counting measure, a mixture of both, or a product measure of the former measures. In particular, this covers both regression and classification tasks or mixtures of the two such as Tobit regression \citep{sigrist2017grabit}. 

The goal of boosting is to find a minimizer $F^*$ of the risk $R(F)$ which is defined as the expected loss
\begin{equation}\label{risk}
R(F)=E_{Y,X}(L(Y,F(X))),
\end{equation}
where $F(\cdot)$ is a function in a Hilbert space $\mathcal{H}$ with inner product $\langle\cdot,\cdot\rangle$ given by
$$\langle F,F\rangle=E_X\left(F(X)^2\right), $$ and $L(Y,F)$ is a loss function. See below and Appendix \ref{lossfcts} for examples of loss functions. For notational simplicity, we often denote a function $F(\cdot)$ shortly by $F$ in this article. In general, $F$ can also be a multivariate function in a direct sum  Hilbert space. However, for notational simplicity, we assume in the following that $F$ is univariate. In Section \ref{mutlivar}, we extend this to the multivariate case.

Boosting assumes that the minimizer $F^* \in \Omega_{\mathcal{S}}$ lies in the span $\Omega_{\mathcal{S}}=span(\mathcal{S})$ of a set $\mathcal{S}$ of base learners $f_j:\mathbb{R}^p\rightarrow \mathbb{R}$:
\begin{equation}\label{minprob}
F^*=\argmin_{F\in \Omega_{\mathcal{S}}}R(F).
\end{equation}
If the risk $R(F)$ is convex in $F$, then \eqref{minprob} is a convex optimization problem since $\Omega_{\mathcal{S}}$ is also convex. Boosting finds $F^*$ in a stagewise way by sequentially adding an update $f_m$ to the current estimate $F_{m-1}$,
\begin{equation}\label{boostupdate}
F_m(x)= F_{m-1}(x)+ f_m(x),~~f_m\in \mathcal{S}, ~~m=1,\dots,M,
\end{equation}
such that the risk is minimized
\begin{equation}\label{minupdt}
f_m=\argmin_{f\in \mathcal{S}}R\left(F_{m-1}+  f\right).
\end{equation} 
This minimization can often not be done analytically and an approximation has to be used. 

Different boosting algorithms vary in the way the minimization in \eqref{minupdt} is done, the loss function $L$ used in \eqref{risk}, and in the choice of base learners $f_j\in \mathcal{S}$. Concerning loss functions, potential choices include the squared loss $L(y,F)=(y-F)^2/2$ for regression, the negative Gaussian log-likelihood where both the mean and the scale parameter depend on predictor variables \citep[see, e.g.,][]{mayr2012generalized}, the negative log-likelihood $-yF+\log\left(1+e^F\right)$ of a binomial model with a logistic link function for binary classification, or the entropy loss with a softmax function for multiclass classification. Under appropriate regularity assumptions, one can use the negative log-likelihood of any statistical model as loss function:
$$
L(y,F)=-\log\left(f_{F,\theta}(y)\right),
$$
where $f_{F,\theta}(y)$ is the density of $Y$ given $X$ with respect to some reference measure, $F$ is linked to one or several, possibly transformed, parameters of this density, and $\theta$ are additional parameters. See Appendix \ref{lossfcts} for various examples of loss functions and, in particular, the ones we consider in the empirical evaluation of this article. As shown by \citet{friedman2000additive}, AdaBoost algorithms are versions of Newton boosting for classification with an exponential loss function.

Concerning base learners, regression trees \citep{breiman1984classification} is the most frequently adopted choice. Other potential base learners include splines or linear functions \citep{buhlmann2003boosting,buehlmann2006boosting,schmid2008boosting}. In this article, we focus on trees:
$$f(x)=w_{s(x)},$$ where $s:\mathbb{R}^p\rightarrow \{1,\dots,J\}$, $w\in \mathbb{R}^J$, and $J\in \mathbb{N}$ denotes the number of terminal nodes, or leaves, of the tree $f(x)$. The function $s$ represents the structure of the tree, i.e., the partition of the space $\mathbb{R}^p$, and $w$ contains the values of the leaves. As in \citet{breiman1984classification}, we assume that the partition of the space made by $s$ is a binary tree where each cell in the partition is a rectangle of the form  $R_j=(l_1,u_1]\times\dots\times(l_p,u_p]\subset\mathbb{R}^p$ with $-\infty\leq l_m<u_m\leq\infty$ and $s(x)=j$ if  $x\in R_j$. 

For finding an update in \eqref{boostupdate}, either a form of gradient descent, Newton's method, or a hybrid variant is used to obtain an approximate solution to the minimization problem in \eqref{minupdt}. In the following, we describe these approaches. 

\subsubsection{Gradient boosting}
Assuming that the risk $R(F)$ is G\^ateau differentiable for all $F\in\Omega_{\mathcal{S}}$, we denote the G\^ateau derivative by
\begin{equation*}
\begin{split}
dR(F,f)&=\frac{d}{d\epsilon}R(F+\epsilon f)\Big|_{\epsilon=0}\\&=\lim_{\epsilon \rightarrow 0}\frac{R(F+\epsilon f)-R(F)}{\epsilon},~~F,f\in\Omega_{\mathcal{S}}.
\end{split}
\end{equation*}
Gradient boosting then works by choosing $f_m$ as the minimizer of a first-order Taylor approximation around $F_{m-1}$ with a penalty on the norm of the base learner:
\begin{equation}\label{1dapprox}
\begin{split}
f_m=&\argmin_{f\in \mathcal{S}}R(F_{m-1})+dR(F_{m-1},f)+\frac{1}{2}\langle f,f\rangle\\=&\argmin_{f\in \mathcal{S}} dR(F_{m-1},f)+\frac{1}{2}\langle f,f\rangle.
\end{split}
\end{equation}
Note that we add the penalty $\frac{1}{2}\langle f,f\rangle$ since the functions $f$ are not necessarily normed and $\langle f,f\rangle$ is not assumed to be constant. 

If we assume that $L(Y,F)$ is differentiable in $F$ for P-almost all $X$ and that the derivative is integrable with respect to the measure of $(Y,X)$, then $dR(F_{m-1},f)$ is given by 
$$dR(F_{m-1},f)=E_{Y,X}\left(g_m(Y,X)f(X)\right),$$
where $g_m(Y,X)$ denotes the gradient of the loss function $L(Y,F)$ with respect to $F$ at the current estimate $F_{m-1}$:
\begin{equation}\label{PopGrad}
g_m(Y,X)=\frac{\partial L(Y,F)}{\partial F}\Big|_{F=F_{m-1}(X)}.
\end{equation}
Consequently, \eqref{1dapprox} can be written as
\begin{equation}\label{1dapproxV2}
\begin{split}
f_m=&\argmin_{f\in \mathcal{S}}E_{Y,X}\left(g_m(Y,X)f(X)+\frac{1}{2} f(X)^2\right)\\
=&\argmin_{f\in \mathcal{S}}E_{Y,X}\left(\left(-g_m(Y,X)- f(X)\right)^2\right).
\end{split}
\end{equation}
This shows that $f_m$ is the $L^2$ approximation to the negative gradient $- g_m(Y,X)$ of the loss function $L(Y,F)$ with respect to $F$ evaluated at the current estimate $F_{m-1}(X)$.

If the following expression is well defined for P-almost all $X$, then the minimization in \eqref{1dapproxV2} can also be done pointwise $$f_m(X) =\argmin_{f\in \mathcal{S}}E_{Y|X}\left(\left(-g_m(Y,X)-f(X)\right)^2\right).$$

\subsubsection{Newton boosting}\label{newtonboostpop}
For Newton boosting, we assume that $R(F)$ is two times G\^ateau differentiable and denote the second G\^ateau derivative by
$$d^2R(F,f)=\frac{d^2}{d\epsilon^2}R(F+\epsilon f)\Big|_{\epsilon=0},~~F,f\in\Omega_{\mathcal{S}}.$$
Newton boosting chooses $f_m$ as the minimizer of a second-order Taylor approximation around $F_{m-1}$:
\begin{equation}\label{2dapprox}
\begin{split}
f_m&=\argmin_{f\in \mathcal{S}}R(F_{m-1})+dR(F_{m-1},f)+\frac{1}{2}d^2R(F_{m-1},f).
\end{split}
\end{equation}
If we assume the P-almost all existence and integrability of the second derivative of $L(Y,F)$ with respect to $F$, then \eqref{2dapprox} can be written as
\begin{equation}\label{2dapproxV2}
\begin{split}
f_m&=\argmin_{f\in \mathcal{S}}E_{Y,X}\left(g_m(Y,X) f(X)+\frac{1}{2}h_m(Y,X) f(X)^2\right)\\
&=\argmin_{f\in \mathcal{S}}E_{Y,X}\left(h_m(Y,X) \left(-\frac{g_m(Y,X)}{h_m(Y,X)} -f(X)\right)^2\right),
\end{split}
\end{equation}
where the gradient $g_m(Y,X)$ is defined in \eqref{PopGrad} and $h_m(Y,X)$ is the second derivative of $L(Y,F)$ with respect to $F$ at $F_{m-1}$:
\begin{equation}\label{HessPop}
h_m(Y,X)=\frac{\partial^2L(Y,F)}{\partial F^2}\Big|_{F=F_{m-1}(X)}.
\end{equation}
The last line in Equation \eqref{2dapproxV2} shows that $f_m$ is the weighted $L^2$ approximation to negative ratio of the gradient over the Hessian $-\frac{g_m(Y,X)}{h_m(Y,X)}$ and the weights corresponds to the second derivative $h_m(Y,X)$. 

If the following expression is well defined for P-almost all $X$, we can again calculate the pointwise minimizer of \eqref{2dapproxV2} as:
\begin{equation*}
f_m(X)=\argmin_{f\in \mathcal{S}}E_{Y|X}\Big(h_m(Y,X) \Big(-\frac{g_m(Y,X)}{h_m(Y,X)} -f(X)\Big)^2\Big).
\end{equation*}

Note that gradient boosting can be seen as a special case of Newton boosting. If the second derivative of the loss function $h_m(Y,X)$ exists and is constant, $h_m(Y,X)=c \in \mathbb{R}\backslash\{0\}$, for P-almost all $X$, then the Newton boosting update in \eqref{2dapproxV2} essentially equals the gradient update in \eqref{1dapproxV2}. Specifically, they are exactly equal if $h_m(Y,X)=1$. Since in practice the update is usually damped, see Equation \eqref{damping} in Section \ref{tunepars}, and the shrinkage parameter $\nu$ is considered a tuning parameter, the two approaches are essentially also equivalent for $h_m(Y,X)=c\neq 1$.

\subsubsection{Hybrid gradient-Newton boosting}\label{hybrid}
A hybrid variant of gradient and Newton boosting proposed in \citet{friedman2001greedy} is obtained by first learning part of the parameters of the base learner using a gradient step and the remaining part using a Newton update. For instance, for trees as bases learners, the structure $s$ of a tree is learned using a gradient update:
$$s_m=\argmin_{s: f=w_s\in \mathcal{S}}E_{Y,X}\left(\left(-g_m(Y,X)- f(X)\right)^2\right),$$
and then, conditional on this, one finds the weights $w$ using a Newton step:
$$w_m=\argmin_{\substack{w:f=w_s\in \mathcal{S}\\s=s_m}}E_{Y,X}\Big(g_m(Y,X) f(X)+\frac{1}{2}h_m(Y,X) f(X)^2\Big).$$

\subsubsection{Line search}
The update step in \eqref{boostupdate} is sometimes presented in the form $F_m(x)= F_{m-1}(x)+ \rho_m f_m(x)$ with $\rho_m\in\mathbb{R}$, where $\rho_m$ is found by doing an additional line-search 
$\rho_m=\argmin_{\rho\in\mathbb{R}} R^e\left(F_{m-1}+\rho f_m\right)$. For gradient boosting,  this has the advantage that the length of the gradient does not depend on the scaling of the loss function. However, we are not considering this approach explicitly here since, first, we assume that the set of base learners $\mathcal{S}$ is rich enough to include not just normalized base learners but base learners of any norm and, second, the line-search often cannot be done analytically and a second-order Taylor approximation is used instead. I.e., the latter case corresponds to a version of hybrid gradient-Newton or Newton boosting.

\subsubsection{Applicability of Newton boosting}\label{applNewt}
As mentioned in Section \ref{newtonboostpop}, there is no difference between the three above presented optimization approaches for loss functions with non-zero and constant second derivatives in $F$. In particular, this holds true for the squared loss function. Further, for loss functions where the second derivative is zero on a non-null set of the support of $X$, such as the least absolute deviation (LAD), any other quantile regression loss function, and the Huber loss, Newton and also hybrid gradient-Newton boosting are not applicable. In these cases, the above-mentioned line-search can be useful in addition to a pure gradient step. Similarly, if a loss function is not P-almost everywhere twice differentiable in $F$, Newton boosting is also not applicable. However, the majority of commonly used loss functions are twice differentiable. 

\subsection{Sample versions}\label{boosttrees}
In the following, we assume that we observe $n$ samples $(y_i,x_i), i=1,\dots,n,$ from the same distribution as the one of $(Y,X)$, and approximate the risk $R(F)$ in \eqref{risk} with the empirical risk $R^e(F)$ obtained by replacing the population distribution with the empirical distribution:
\begin{equation}\label{emprisk}
R^e(F)=\frac{1}{n}\sum_{i=1}^n L(y_i,F(x_i)).
\end{equation}

For gradient boosting, the sample version of \eqref{1dapproxV2} can be written as
\begin{equation}\label{1dapproxfin}
\begin{split}
f_m=&\argmin_{f\in \mathcal{S}}\sum_{i=1}^n g_{m,i} f(x_i)+\frac{1}{2}f(x_i)^2\\
=&\argmin_{f\in \mathcal{S}}\sum_{i=1}^n \left(-g_{m,i}-f(x_i)\right)^2,
\end{split}
\end{equation}
where $g_{m,i}$ is the gradient of the loss function for observation $i$
$$g_{m,i}=\frac{\partial}{\partial F}L(y_i,F)\Big|_{F=F_{m-1}(x_i)}. $$
This means that the stagewise minimizer $f_m$ can be found as the least squares approximation to the negative gradient $-g_{m,i}$.

Similarly, the sample version of the Newton update in \eqref{2dapproxV2} is given by
\begin{equation}\label{2dapproxfin}
\begin{split}
f_m&=\argmin_{f\in \mathcal{S}}\sum_{i=1}^n g_{m,i} f(x_i)+h_{m,i}\frac{1}{2} f(x_i)^2\\
&=\argmin_{f\in \mathcal{S}}\sum_{i=1}^n h_{m,i}\left(-\frac{g_{m,i}}{h_{m,i}}-f(x_i)\right)^2,
\end{split}
\end{equation}
where $h_{m,i}$ is the Hessian of the loss function for observation $i$:
\begin{equation}\label{hessdata}
h_{m,i}=\frac{\partial^2}{\partial F^2}L(y_i,F)\Big|_{F=F_{m-1}(x_i)}.
\end{equation}
I.e., $f_m$ can be found as the weighted least squares approximation to the ratio of the negative gradient over the Hessian $-\frac{g_{m,i}}{h_{m,i}}$ with weights given by $h_{m,i}$.

The sample version of the hybrid gradient-Newton algorithm first finds the structure $s$ of a tree using a gradient step:
$$s_m=\argmin_{s: f=w_s\in \mathcal{S}}\sum_{i=1}^n \left(-g_{m,i}-f(x_i)\right)^2,$$
and then determines the weights $w$ using a Newton step:
$$w_m=\argmin_{w:f=w_s\in \mathcal{S},s=s_m}\sum_{i=1}^n h_{m,i}\left(-\frac{g_{m,i}}{h_{m,i}}-f(x_i)\right)^2.$$

\subsection{Multivariate case}\label{mutlivar}
In this section, we briefly present gradient and Newton boosting when the function $\vect{F}$ is multivariate. In this case, $$\vect{F}(X)=(F^1(X),F^2(X),\dots,F^d(X))^T$$ is assumed to be a function in a direct sum Hilbert space $\mathcal{H}=\mathcal{H}^1\oplus\mathcal{H}^2\oplus\dots \oplus\mathcal{H}^d,$ where the $\mathcal{H}^k$'s are Hilbert spaces with inner products $\langle\cdot,\cdot\rangle_k$ given by
$\langle F^k,F^k\rangle_k=E_X\left(F^k(X)^2\right), $
and the inner product for $\mathcal{H}$ is given by
$\langle \vect{F},\vect{F}\rangle=\sum_{k=1}^d \langle F^k,F^k\rangle_k. $
For the sake of readability, we use boldface in this subsection to distinguish vector-valued functions from scalar-valued functions. Examples of loss functions where $\vect{F}$ is multivariate include the entropy loss with a softmax function for multiclass classification or generalized additive models for location, scale, and shape (GAMLSS) where location, scale, and shape parameters are modeled as functions of predictor variables $X$ \citep{rigby2005generalized,mayr2012generalized}.

A gradient boosting update $\vect{f_m}\in \mathcal{S}\oplus\dots \oplus\mathcal{S}$ in Equation \eqref{boostupdate} is then obtained as
\begin{equation}\label{1dapproxMult}
\vect{f_m}=\argmin_{f\in \mathcal{S}\oplus\dots \oplus\mathcal{S}} dR(\vect{F_{m-1}},\vect{f})+\frac{1}{2}\langle \vect{f},\vect{f}\rangle.
\end{equation}
Under appropriate regularity conditions, $dR(\vect{F_{m-1}},\vect{f})$ is given by
$$dR(\vect{F_{m-1}},\vect{f})=E_{Y,X}\left(\vect{g_m}(Y,X)^T\vect{f}(X)\right),$$
where
$$ \vect{g_m}(Y,X)=\left(\frac{\partial}{\partial F^1}L(Y,\vect{F}),\dots,\frac{\partial}{\partial F^d}L(Y,\vect{F})\right)^T\Big|_{\vect{F}=\vect{F_{m-1}}(X)}.$$

It follows that the solution in \eqref{1dapproxMult} can be determined for each $k$, $k=1,\dots,d$, separately as
\begin{equation*}
f^k_m=\argmin_{f^k\in \mathcal{S}}E_{Y,X}\left(\left(-g^k_m(Y,X)- f^k(X)\right)^2\right),
\end{equation*}
where $$g^k_m(Y,X)=\frac{\partial}{\partial F^k}L(Y,\vect{F})\Big|_{\vect{F}=\vect{F_{m-1}}(X)}.$$ 
The sample version of this gradient boosting update is given by
\begin{equation*}
f^k_m=\argmin_{f^k\in \mathcal{S}}\sum_{i=1}^n \left(-g^k_{m,i}-f^k(x_i)\right)^2,
\end{equation*}
where $g^k_{m,i}=g^k_m(y_i,x_i). $

Newton boosting obtains an update $\vect{f_m}\in \mathcal{S}\oplus\dots \oplus\mathcal{S}$ as
\begin{equation*}
\vect{f_m}=\argmin_{f\in \mathcal{S}\oplus\dots \oplus\mathcal{S}}dR(\vect{F_{m-1}},\vect{f})+\frac{1}{2}d^2R(\vect{F_{m-1}},\vect{f}),
\end{equation*}
where, again under appropriate conditions, this can also be written as
\begin{equation*}
f_m=\argmin_{f\in \mathcal{S}\oplus\dots \oplus\mathcal{S}}E_{Y,X}\left(\vect{g_m}(Y,X)^T \vect{f}(X)+\frac{1}{2}\vect{f}(X)^T\vect{h_m}(Y,X) \vect{f}(X)\right)
\end{equation*}
with $\vect{h_m}(Y,X)=\left[h_m(Y,X)\right]_{k,l}$, $k,l=1,\dots,d$, and
\begin{equation*}
\left[h_m(Y,X)\right]_{k,l}=\frac{\partial^2}{\partial F^k\partial F^l}L(Y,\vect{F})\Big|_{\vect{F}=\vect{F_{m-1}}(X)}.
\end{equation*}

The sample version of the Newton update is given by
$$
\vect{f_m}=\argmin_{f\in \mathcal{S}\oplus\dots \oplus\mathcal{S}}\sum_{i=1}^n \vect{g_{m,i}}^T \vect{f}(x_i)+\frac{1}{2}\vect{f}(x_i)^T\vect{h_{m,i}} \vect{f}(x_i),
$$
where $\vect{g_{m,i}}=\vect{g_m}(y_i,x_i) $ and $\vect{h_{m,i}}=\vect{h_m}(y_i,x_i)$. In practice, one often approximates $\vect{h_{m,i}}$ by a diagonal matrix 
$$\vect{h_{m,i}}\approx\text{diag}\left(\frac{\partial^2}{{\partial F^k}^2}L(y_i,\vect{F})\Big|_{\vect{F}=\vect{F_{m-1}}(x_i)}\right).$$ In this case, the updates can be determined independently as
\begin{equation*}
f^k_m=\argmin_{f^k\in \mathcal{S}}\sum_{i=1}^n h^k_{m,i}\left(-\frac{g^k_{m,i}}{h^k_{m,i}}-f^k(x_i)\right)^2,
\end{equation*}
where $h^k_{m,i}=h^k_{m}(y_i,x_i)$.

\subsection{Tuning parameters and regularization}\label{tunepars}
It has been empirically observed that damping the update in \eqref{boostupdate} results in increased predictive accuracy \citep{friedman2001greedy}. This means that the update in \eqref{boostupdate} is replaced with
\begin{equation}\label{damping}
F_m(x)= F_{m-1}(x)+ \nu f_m(x),~~\nu>0,
\end{equation}
where $\nu$ is a shrinkage parameter or learning rate. The parameter $\nu$ can be thought of as a regularization parameter. Under additional assumptions, one can show for linear base learners that when the parameter $\nu$ goes to zero, the obtained solutions correspond to the set of Lasso solutions \citep{efron2004least,zhao2007stagewise}.

The main tuning parameters of boosting algorithms are thus the number of boosting iterations $M$ and the shrinkage parameter $\nu$. These tuning parameters and also the ones for the base learners presented in the following can be chosen by minimizing a performance measure on a validation dataset, using cross-validation, or using an appropriate model selection criterion. 

\subsubsection{The minimum equivalent sample size per leaf parameter}\label{novelpar}
Depending on the choice of base learners, there are additional tuning parameters. For instance, if trees are used as base learners, the depth of the trees $L$ and the minimum number of samples per leaf are tuning parameters. Since Newton boosting solves the weighted least squares problem in \eqref{2dapproxfin} in each update step, the raw number of samples per leaf is not meaningful, and we argue that instead, one should consider what we denote as the \textit{equivalent sample size per leaf} per leaf. As we show below on real-world and simulated data, this parameter can be important for predictive accuracy.

Specifically, we first normalize the weights
$$\tilde w_{m,i}= n\cdot \frac{h_{m,i}}{\sum_{j=1}^n h_{m,j}},$$
such that the sum of all normalized weights $\tilde w_{m,i}$ equals the number of data points $n$. We then denote the sum of all normalized weights $\sum_{i\in L_j} \tilde w_{m,i}$ per leaf $L_j$ as the equivalent sample size per leaf, or equivalent number of weighted data points, and require that this is larger than a certain constant $S$:
\begin{equation}\label{minsamp}
\sum_{i\in L_j} \tilde w_{m,i}\geq S.
\end{equation} 
The constant $S$ is considered as a tuning parameter analogous to the minimum sample size per leaf in gradient boosting.

To the best of our knowledge, other software implementations that use Newton boosting such as \texttt{XGBoost} \citep{chen2016xgboost} and \texttt{LightGBM} \citep{ke2017lightgbm} handle this tuning parameter differently by requiring that the sum of all raw weights $h_{m,i}$ per leaf is larger than a certain constant which is by default one.\footnote{This constant is denoted by \texttt{min\_child\_weight} in \texttt{XGBoost} (as of September 10, 2020).} According to the authors of \texttt{XGBoost}, the motivation for this is that for linear regression, ``this simply corresponds to minimum number of instances needed to be in each node''.\footnote{Unfortunately, this is not documented in the corresponding companion article \citep{chen2016xgboost} We gather this information from the online documentation https://xgboost.readthedocs.io/en/latest/parameter.html (retrieved on September 10, 2020).} We argue that this is not a good choice for the following reasons.

First, the second derivative $h_{m,i}$ of the loss function of a linear regression model with Gaussian noise $L(Y,F)=\frac{(Y-F)^2}{2\sigma^2}$ equals one only if the noise variance $\sigma^2$ equals one $\sigma^2=1$. Otherwise, the second derivative $h_{m,i}$ equals $\sigma^{-2}$. This means that the analogy to the linear regression case does not hold true in general. In contrast, our proposed normalized weights $\tilde w_{m,i}$ do indeed equal one for the linear regression case no matter what the noise variance is, and thus the sum of normalized weights $\sum_{i\in L_j} \tilde w_{m,i}$ equals the number of samples per leaf for the linear regression model also when $\sigma^2\neq 1$. In general, the sum of normalized weights $\tilde w_{m,i}$ corresponds to the number of weighted samples, and one has thus good intuition concerning reasonable candidate values or ranges for this. If the raw weights are not normalized, this is not the case. I.e., the sum of raw weights cannot be interpreted as the number of weighted samples, and its interpretation changes depending on the loss function used. Consequently, the minimum sum of raw weights $\sum_{i\in L_j} h_{m,i}$ is a parameter that is difficult to tune in practice and we obtain inferior predictive accuracy for the large majority of datasets in our experiments in Section \ref{realdata}. Further, in Section \ref{notunemain} we provide empirical evidence that the minimum number of weighted samples per leaf is an important tuning parameter and that the unnormalized version of this tuning parameter is difficult to tune.

In addition to the above-presented tuning parameters, one can consider further tuning parameters such as $L^1$ and/or $L^2$ regularization penalties on the tree weights, or an $L^0$ penalty on the number of leaves. Finally, boosting algorithms can also be made stochastic \citep{friedman2002stochastic} by (sub-)sampling data points in each boosting iteration and variables in the tree algorithm as it is done for random forests. 

\subsection{Numerical stability and computational cost}
\citet{friedman2000additive} observed for the LogitBoost algorithm, i.e. Newton boosting for a Bernoulli likelihood with a logistic link function, that numerical stability can be an issue for Newton boosting. Similarly as in \citet{friedman2000additive}, we enforce a lower bound on the second derivatives $h_{m,i}$ at $10^{-20}$ such that they are always strictly positive in our implementation of Newton boosting.\footnote{We have not done a comprehensive study on the impact of this lower bound. However, when we choose the bounds at $10^{-16}$ and $10^{-30}$, we have not observed any noticeable differences in the outcomes (results not tabulated).}

Concerning computational cost, the main cost of a boosting algorithm with trees as base learners results from growing the regression trees \citep{ke2017lightgbm}. Consequently, the differences in computational times are marginal for the three versions of boosting presented in this article. Tree boosting implementations that are designed to scale to large data use computational efficient algorithms for growing trees; see, e.g., \citet{chen2016xgboost}.

\subsection{Software implementations}
The methodology presented in this article, i.e., gradient, Newton, and hybrid gradient-Newton boosting is implemented in the Python package \texttt{KTBoost}, which is openly available from the Python Package Index (PyPI) repository.\footnote{The parameter \texttt{update\_step} of the functions \texttt{BoostingClassifier} and \texttt{BoostingRegressor} takes as arguments \texttt{gradient}, \texttt{hybrid}, or \texttt{newton}. See https://github.com/fabsig/KTBoost for more information.}

We briefly summarize which types of boosting algorithms are used by existing software implementations. The R package \texttt{gbm} \citep{ridgeway2007generalized} and the Python library \texttt{scikit-learn} \citep{scikit-learn} follow the approach of \citet{friedman2001greedy} and use gradient descent steps for finding the structures of trees with Newton updates for the tree leaves (if applicable, see Section \ref{applNewt}). \texttt{XGBoost} \citep{chen2016xgboost} uses Newton boosting with Newton steps for finding both the tree structure and the tree leaves. The R package \texttt{mboost} \citep{hothorn2010model} uses gradient boosting. In addition to trees, it also supports other base learners which include linear functions, one- and two-dimensional smoothing splines, spatial terms, as well as user-defined ones. Other recent implementations such as \texttt{LightGBM} \citep{ke2017lightgbm}, \texttt{TF Boosted Trees} \citep{ponomareva2017tf} and \texttt{Spark MLLib} \citep{meng2016mllib}, do not explicitly mention in their companion articles \citep{ke2017lightgbm, ponomareva2017tf} or in their online documentation\footnote{https://spark.apache.org/docs/latest/mllib-ensembles.html\#gradient-boosted-trees-gbts (retrieved on September 10, 2020).} whether gradient descent or Newton updates are used in the stagewise boosting updates. We infer from the corresponding source code that \texttt{LightGBM} uses Newton boosting. To the best of our knowledge, none of the existing solutions allows the user to explicitly choose between a gradient or a Newton step for calculating the boosting updates. 

\section{Empirical evaluation and comparison}\label{realdata}
In the following, we compare the three different boosting algorithms presented in the previous section for different loss functions on various datasets using regression trees as base learners.\footnote{The code to reproduce the results can be found on https://github.com/fabsig/GradientNewtonBoosting} Specifically, we use the CART version of \citet{breiman1984classification} with the mean squared error as splitting criterion. Note that we use trees \citep{breiman1984classification} as base learners as these are the most widely adopted base learners in applied data science and machine learning research \citep{ridgeway2007generalized,scikit-learn,chen2016xgboost,meng2016mllib,ke2017lightgbm,ponomareva2017tf}. Besides Newton boosting with the novel equivalent sample size per leaf parameter, we also consider Newton boosting as implemented in \texttt{XGBoost} for which the sum of Hessians in each leaf acts as tuning parameter.\footnote{We use \texttt{XGBoost} version number 0.7 in Python with the options \texttt{tree\_method=`exact'}, \texttt{updater=`grow\_colmaker'}, \texttt{lambda=0}, and all other parameters at the default values unless otherwise mentioned.}

\subsection{Real-world data}
We consider the following datasets: adult, bank, (breast) cancer, ijcnn, ionosphere, titanic, sonar, car, covtype, digits, glass, letter, satimage, smartphone, usps, insurance, birthweight, and (childhood) malnutrition. Poisson regression is used for the insurance dataset. For the birthweight and malnutrition datasets, we use mean-scale regression assuming a normal likelihood where both the mean and the log-transformed scale parameter, i.e. the log-transformed standard deviation, are modeled as functions of the predictor variables; see Appendix \ref{lossfcts} for more details. Note that the mean-scale regression model is an example of a GAMLSS model \citep{rigby2005generalized,mayr2012generalized}. For the remaining datasets, binary or multiclass classification is used. The insurance dataset is obtained from Kaggle\footnote{https://www.kaggle.com/apex51/poisson-regression}. The birthweight \citep{schild2008weight} and malnutrition \citep{fenske2011identifying} datasets are obtained from the \texttt{tbm} R package\footnote{Available on https://r-forge.r-project.org}. The covtype, ijcnn, and usps datasets are LIBSVM datasets\footnote{https://www.csie.ntu.edu.tw/\textasciitilde{}cjlin/libsvmtools/datasets/}. All other datasets are obtained from the UCI Machine Learning Repository\footnote{http://archive.ics.uci.edu/ml/datasets/}. A summary of the datasets can be found in Table \ref{data_sum}. If a dataset contains categorical predictor variables, these are converted to binary dummy variables using one-hot encoding.

\begin{table}[ht!]
\centering
\begin{tabular}{llrr}
  \hline
\hline
Data & Type / nb. classes & Nb. samples & Nb. features \\ 
  \hline
adult & 2 & 48842 & 108 \\ 
  bank & 2 & 41188 &  62 \\ 
  cancer & 2 & 699 &   9 \\ 
  ijcnn & 2 & 141691 &  22 \\ 
  ionosphere & 2 & 351 &  34 \\ 
  sonar & 2 & 208 &  60 \\ 
  car & 4 & 1728 &  21 \\ 
  covtype & 7 & 581012 &  54 \\ 
  digits & 10 & 5620 &  64 \\ 
  glass & 7 & 214 &   9 \\ 
  letter & 26 & 20000 &  16 \\ 
  satimage & 6 & 6438 &  36 \\ 
  smartphone & 6 & 10299 & 561 \\ 
  usps & 10 & 9298 & 256 \\ 
   \hline
insurance & Poisson regr. & 50999 & 117 \\ 
  birthweight & Mean-scale regr. & 150 &   5 \\ 
  malnutrition & Mean-scale regr. & 24166 &  42 \\ 
   \hline
\hline
\end{tabular}
\caption{Summary of datasets.} 
\label{data_sum}
\end{table}

We randomly split the data into three equally sized datasets: training, validation, and test data. Learning is done on the training data, tuning parameters are chosen on the validation data, and model comparison is done on the test data. For the two largest datasets (ijcnn and covtype) we limit the size of the training, validation, and test data to 20000 data points. This is done for computational reasons. We note that there are various strategies so that tree-based boosting scales to large data \citep{chen2016xgboost,ke2017lightgbm}, but this is not the scope of this article. To quantify variability in the results, we use several different random splits of the data. The number of sample splits is 100 for datasets with less than 1500 samples (less than 500 training samples), 20 for datasets with a size between 1500 and 7500 (number of training samples between 500 and 2500), and 10 for datasets with more than 7500 samples (more than 2500 training samples).

Concerning tuning parameters, we select the number of boosting iterations $M$ from $\{1,2,\dots, 1000\}$, the learning rate $\nu$ from  $\{1,10^{-1},10^{-2},10^{-3}\}$, and the minimum number of samples per leaf from $\{1,5,25,100\}$. For Newton boosting, the latter is replaced by the equivalent sample size per leaf in Equation \eqref{minsamp}, and for the \texttt{XGBoost} implementation, the minimum sum of Hessians per leaf (\texttt{min\_child\_weight}) is used. Further, for the mean-scale regression datasets, the minimum number of samples per leaf is chosen from $\{25,100\}$ only for gradient and hybrid boosting since a very small number of samples can lead to identifiability problems when modeling both the mean and the scale. Tuning parameters are chosen for each sample split such that they minimize the error rate for classification and the negative log-likelihood for regression on the validation data. The maximal tree depth is set to five for all methods. We are not considering the maximal tree depth as an additional tuning parameter for computational reasons.  However, additional results for a subset of the datasets reported in Section \ref{maxtreedepth} and Appendix \ref{treedepth} show that similar findings are obtained for other tree depths. Further, we note that in Section \ref{notunemain} and Appendix \ref{notune}, we also consider the case when the minimum number of samples per leaf parameter is not chosen by minimizing the test error on the validation data but is instead set to the default value of one.

\begin{figure}[ht!]%
	\centering
	\includegraphics[width=0.9\textwidth]{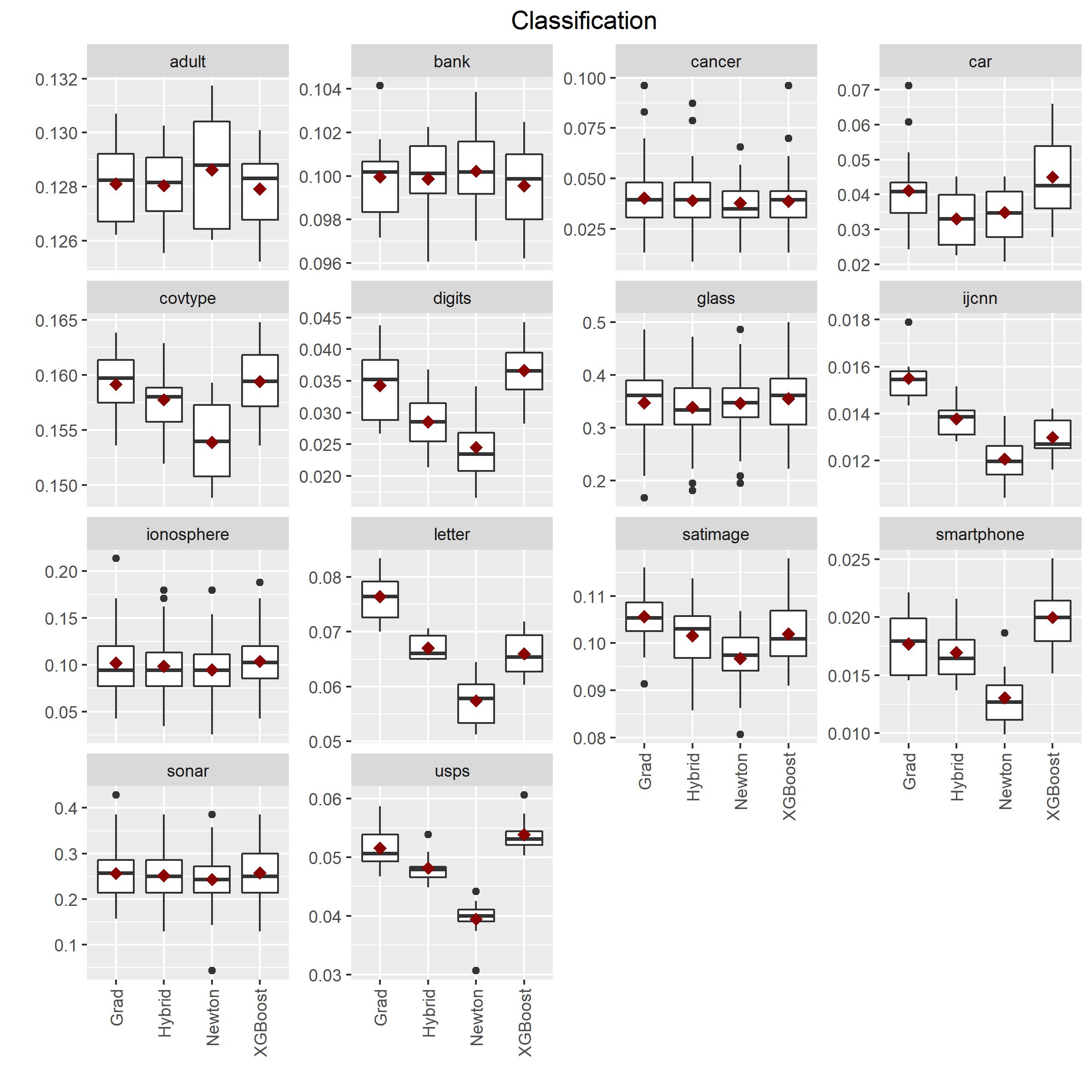}
	\includegraphics[width=0.65\textwidth,trim={1cm 0.5cm 0 0.5cm}]{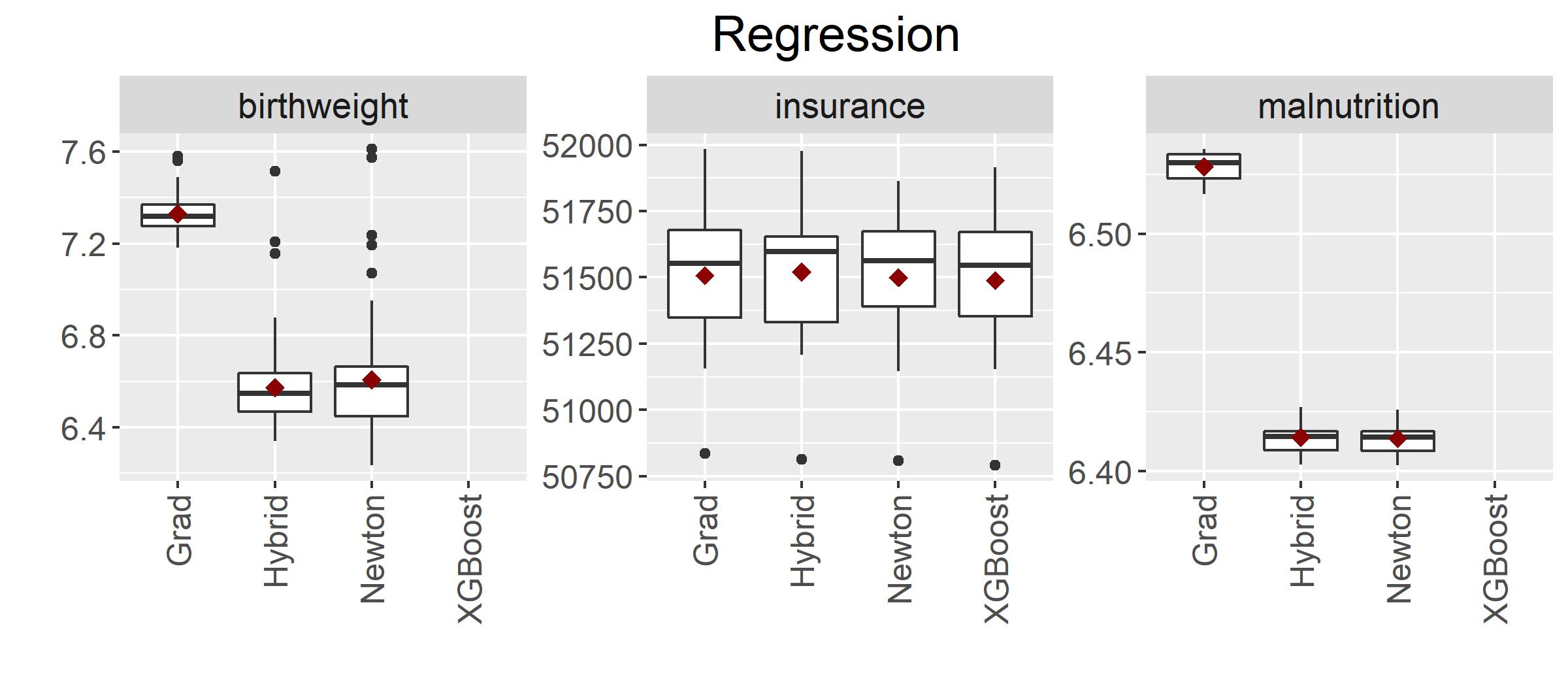}
	\caption{Comparison of boosting methods using test error rate for classification and test negative log-likelihood for regression. The red rhombi represent means.}
	\label{real_data_score} 
\end{figure}

\begin{table}[ht!]
\centering
\begingroup\footnotesize
\begin{tabular}{lllll}
  \hline
\hline
Data & Grad & Hybrid & Newton & XGBoost \\ 
  \hline
adult & 0.128 (0.00158) & 0.128 (0.00143) & 0.129 (0.00222) & \textbf{0.128} (0.00158) \\ 
  bank & 0.1 (0.00213) & 0.0999 (0.00191) & 0.1 (0.00208) & \textbf{0.0996} (0.00198) \\ 
  cancer & 0.0402 (0.0147) & 0.039 (0.0132) & \textbf{0.0378} (0.0108) & 0.0387 (0.0124) \\ 
  ijcnn & 0.0155 (0.00101) & 0.0138 (0.000791) & \textbf{0.0121} (0.00109) & 0.013 (0.000838) \\ 
  ionosphere & 0.102 (0.0317) & 0.0984 (0.0293) & \textbf{0.0945} (0.0277) & 0.104 (0.0288) \\ 
  sonar & 0.256 (0.0556) & 0.252 (0.0517) & \textbf{0.243} (0.0499) & 0.257 (0.055) \\ 
  car & 0.0411 (0.0114) & \textbf{0.0331} (0.00759) & 0.0349 (0.00764) & 0.045 (0.0112) \\ 
  covtype & 0.159 (0.00331) & 0.158 (0.00326) & \textbf{0.154} (0.00384) & 0.159 (0.00371) \\ 
  digits & 0.0343 (0.00547) & 0.0285 (0.00464) & \textbf{0.0245} (0.00467) & 0.0367 (0.0044) \\ 
  glass & 0.347 (0.0618) & \textbf{0.339} (0.0612) & 0.346 (0.0595) & 0.355 (0.0607) \\ 
  letter & 0.0764 (0.00449) & 0.067 (0.00233) & \textbf{0.0574} (0.00438) & 0.066 (0.00409) \\ 
  satimage & 0.106 (0.00622) & 0.102 (0.00801) & \textbf{0.0968} (0.00673) & 0.102 (0.00687) \\ 
  smartphone & 0.0177 (0.00272) & 0.017 (0.00266) & \textbf{0.013} (0.00267) & 0.02 (0.00288) \\ 
  usps & 0.0516 (0.00371) & 0.0482 (0.00261) & \textbf{0.0395} (0.00364) & 0.0539 (0.00311) \\ 
   \hline
insurance & 51500 (341) & 51500 (349) & 51500 (324) & \textbf{51500} (332) \\ 
  malnutrition & 6.53 (0.00649) & 6.41 (0.00792) & \textbf{6.41} (0.00733) &  \\ 
  birthweight & 7.33 (0.0702) & \textbf{6.57} (0.177) & 6.61 (0.226) &  \\ 
   \hline
Av. rank & 3.27 & 2.2 & 1.6 & 2.93 \\ 
   \hline
p-val Friedman test & 0.000746 &  &  &  \\ 
  Adj. p-val Wilcoxon test & 0.000229 & 0.0714 &  & 0.0302 \\ 
   \hline
\hline
\end{tabular}
\endgroup
\caption{Results for real-world data: Average test error rates for classification and test negative log-likelihoods for regression. 
    In parentheses are approximate standard deviations. 
    Below are average ranks of the methods over the different datasets (only considering datasets for which all four methods are run). 
    Further, a p-value of a Friedman test with an Iman and Davenport correction for comparing the different algorithms is reported.
    The last row shows Holm-Bonferroni corrected p-values of Wilcoxon signed-rank tests for pairwise 
    comparison of Newton boosting with the novel number of weighted samples parameter and the three alternative methods.} 
\label{results_realdata}
\end{table}

In Figure \ref{real_data_score} and Table \ref{results_realdata}, we report test error rates for classification and test negative log-likelihoods for regression datasets. Figure \ref{real_data_score} visualizes the results using boxplots. In Table \ref{results_realdata}, we additionally report average test errors and test negative log-likelihoods as well as approximate standard deviations. Further, we report the average rank of every method over the different datasets. Since \texttt{XGBoost} does not support mean-scale regression, we only consider the datasets for which all four methods can be run when calculating average ranks. Overall, we find that Newton boosting with the novel equivalent sample size per leaf parameter has clearly the lowest generalization error among the four methods. Its average rank i1 $1.6$. The second best method with an average rank of $2.2$ is hybrid gradient-Newton boosting. Gradient boosting often has the lowest predictive accuracy with an average rank of $3.27$. In addition, Newton boosting with the novel number of weighted samples parameter performs substantially better than the \texttt{XGBoost} variant of Newton boosting with a minimum sum of unnormalized Hessians parameter which has an average rank of $2.93$. We observe particularly striking differences with large outperformance in the predictive accuracy of Newton boosting for several classification datasets (ijcnn, digits, letter, satimage, smartphone, and usps). For the two mean-scale regression datasets (birthweight and malnutrition), we also observe that gradient boosting performs worse than Newton and hybrid gradient-Newton boosting, but no notable difference among the latter two is found. For the Poisson regression dataset (insurance), gradient, hybrid, and Newton boosting perform equally well. 

Concerning statistical significance, we note that when using a resampling approach, standard statistical tests, such as a paired t-test, cannot be used to do a pairwise comparison of the different algorithms separately per dataset since training and test datasets in different splits are dependent due to overlap \citep{dietterich1998approximate, bengio2004no, demvsar2006statistical}, and this can result in biased standard error estimates for the generalization error. Following \citet{demvsar2006statistical}, we compare the different methods across all datasets using a Friedman test with an Iman and Davenport correction \citep{iman1980approximations}. This gives a p-value of $0.000746$ which shows that the differences in the four methods are highly significant. We next use a Wilcoxon signed-rank test to investigate whether the pairwise differences in accuracy between Newton boosting with the novel number of weighted samples parameter and the three alternative methods are statistically significant. To account for the fact that we do multiple tests, we apply a Holm-Bonferroni correction \citep{holm1979simple}. Comparing Newton boosting with gradient and hybrid gradient-Newton boosting, we obtain adjusted p-values of $0.000229$ and $0.074$. I.e., Newton boosting performs significantly better than gradient boosting and the difference between Newton boosting and hybrid gradient-Newton boosting is marginally not significant at a $5\%$ level. However, the sample size for performing these tests is relatively small ($17$) and, consequently, the tests likely have low power.

\subsection{Simulated data}\label{simdata}
In the following, we compare the performance of the different boosting approaches on simulated data for both classification and regression. Concerning regression, we consider two extensions of generalized linear models, boosted Poisson and Gamma regression, as well as the mean-scale regression model used in Section \ref{realdata}. For classification, we consider both binary and multiclass classification. In addition, we consider the boosted Tobit model \citep{sigrist2017grabit}, which can be interpreted as a hybrid regression-classification model. See Section \ref{lossfcts} in the appendix for more details on these models.

For classification, we use the \texttt{scikit-learn} function \texttt{make\_classification}, which simulates from an algorithm that is adapted from \citet{guyon2003design} and was designed to generate the `Madelon' dataset. We use this for both simulating binary data and a multiclass data with five classes. Further, we assume ten (informative) features and no redundant and repeated features; see \citet{guyon2003design} for more details. These two datasets are denoted by `bin\_classif' and `multi\_classif' in the following. In addition, we simulate binary data according to the following specification introduced in \citet{friedman2000additive}:
\begin{equation*}
\begin{split}
F(X)&=10\sum_{j=1}^{6}X_j\left(1+\sum_{l=1}^{6}(-1^{l})X_l\right),~~~~X\sim N(0,I_{10}),\\
Y|X&\sim \text{Bernoulli}(p), ~~p=\left(1+e^{-F(X)}\right)^{-1}.
\end{split}
\end{equation*}
This data is denoted by `bin\_classif\_fht' in the following. Finally, we also simulate multiclass data with five classes according to the following specification \citep{friedman2000additive}:
\begin{equation*}
\begin{split}
R^2&=\sum_{j=1}^{10}X_j^2,~~~~X\sim N(0,I_{10}),\\
Y&=k~~ \text{ if }~~ t_k\leq R^2<t_{k+1},\\
\end{split}
\end{equation*}
where the thresholds $t_k$ are chosen such that the labels are approximately equally distributed among the different classes. We denote this data by `multi\_classif\_fht'.

For Poisson and Gamma regression, the boosted Tobit model, as well as mean-scale regression ('msr'), we consider two non-linear functions. First, we use a function of \citet{friedman1991multivariate} given by:
$$F(X)=5\cdot \tan^{-1}\left(\frac{X_2X_3-1-\frac{1}{X_2X_4}}{X_1}\right)+0.2 ~~(\text{`\_f3'}),$$
where $X=(X_1,X_2,X_3,X_4)'$ with $X_1\sim Unif(0,100)$, $X_2\sim Unif(40\pi,560\pi)$, $X_3\sim Unif(0,1)$, and $X_4\sim Unif(1,11).$ In contrast to the original function of \citet{friedman1991multivariate}, we multiply the function by $5$ and add $0.2$ such that function also attains larger values and that all values are positive. We use the \texttt{scikit-learn} function \texttt{make\_friedman3} for simulation and denote datasets generated by this function using the suffix `\_f3' in the following. Further, we also consider a function introduced in \citet{ridgeway1999generalization}:
\begin{equation*}
F(X)=\exp\big(2\sin(3X_1+5X_1^2)-2\sin(3(X_2+0.1)+5(X_2+0.1)^2)\big) ~~(\text{`\_r'}),
\end{equation*}
where $X=(X_1,X_{2})'$, $X_j\sim Unif(0,1)$, independent. Datasets generated using this function are denoted by the suffix `\_r'.\footnote{For instance, the dataset 'msr\_r' is simulated from a mean-scale regression model with both the mean and the standard deviation given by the above function introduced by \citet{ridgeway1999generalization}.} For Poisson and Gamma regressions, the above functions are used to model the mean, and for Tobit regression the functions model the mean of the latent variable. For the mean-scale regression model, we simulate $2n$ variables and relate both the mean and the logarithmic standard deviation to half of the variables. Both Tobit regression and a regression model where both the mean and the scale depend on predictor variables are not supported in \texttt{XGBoost} and, consequently, no comparison can be done for these. For Gamma regression, we set the shape parameter to $\gamma=10$ and consider this as a known parameter.\footnote{We note that \texttt{XGBoost} only supports Gamma regression for $\gamma=1$. However, this slight miss-specification seems to have no detrimental impact as our results below show.} For the Tobit model, we use $\sigma=1$ and also consider this as a known parameter. Further, we set the lower and upper censoring thresholds $y_l$ and $y_u$ in such a way that approximately one-third of all data points are lower and upper censored.

We simulate 10 times datasets with 15000 samples. In each run, 5000 samples are used as training, validation, and test data. As in Section \ref{realdata}, we calculate the p-value of a Friedman test with an Iman and Davenport correction \citep{iman1980approximations} to check whether there are significant differences among the methods across all datasets. Further, we calculate Holm-Bonferroni corrected \citep{holm1979simple} p-values of Wilcoxon signed-rank tests for pairwise comparison of Newton boosting with the three other approaches.  

\begin{figure}[ht!]
	\centering
	\includegraphics[width=0.9\textwidth]{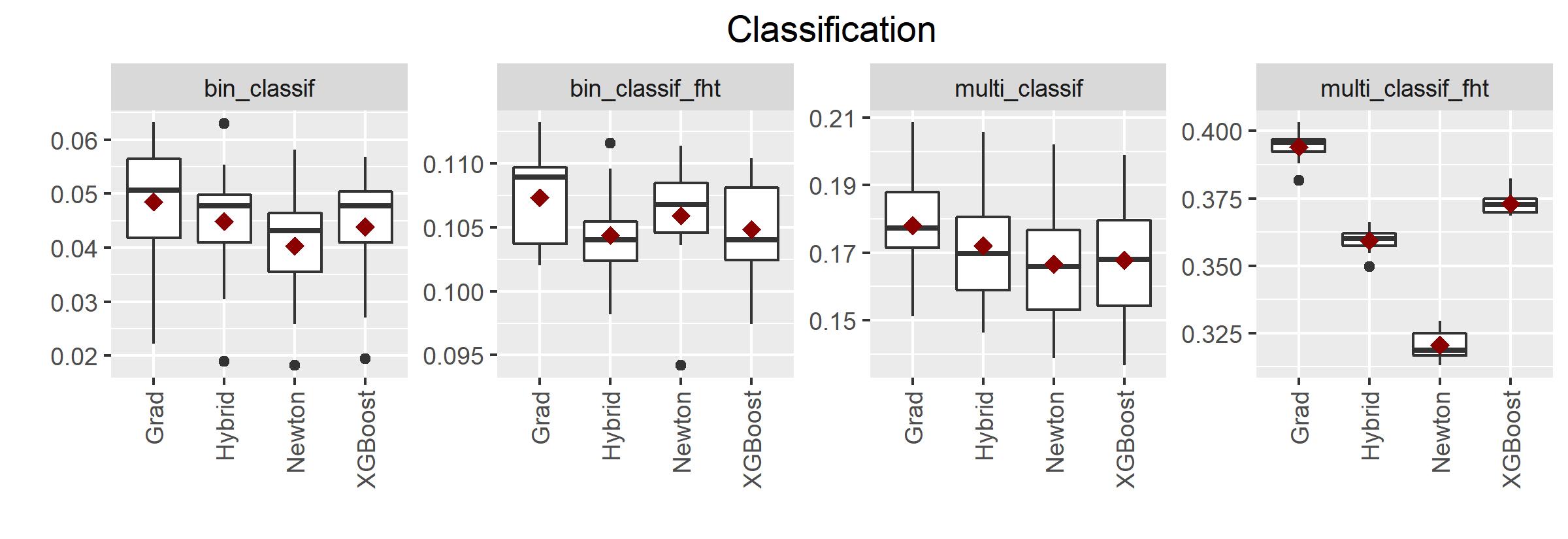}
	\includegraphics[width=0.9\textwidth]{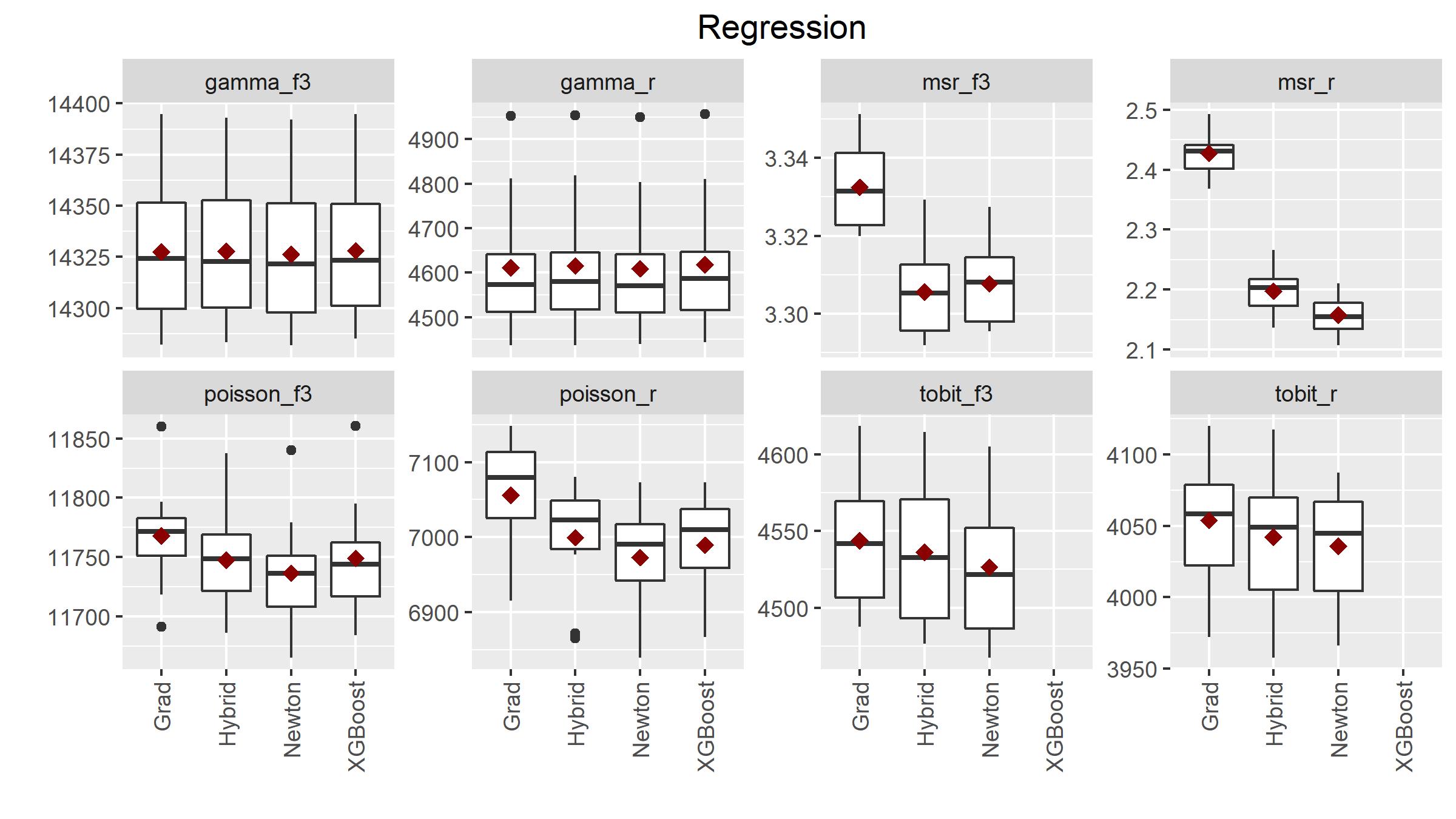}
	\caption{Comparison of boosting methods on simulated datasets using test error rate for classification and test negative log-likelihood for regression. The red rhombi represent means.}
	\label{sim_data_score} 
\end{figure}

\begin{table}[ht!]
\centering
\begingroup\footnotesize
\begin{tabular}{lllll}
  \hline
\hline
Data & Grad & Hybrid & Newton & XGBoost \\ 
  \hline
bin\_classif & 0.0485 (0.0126) & 0.0449 (0.0126) & \textbf{0.0403} (0.0116) & 0.0438 (0.0121) \\ 
  bin\_classif\_fht & 0.107 (0.00394) & \textbf{0.104} (0.004) & 0.106 (0.00488) & 0.105 (0.00428) \\ 
  multi\_classif & 0.178 (0.0163) & 0.172 (0.0176) & \textbf{0.167} (0.0184) & 0.168 (0.0185) \\ 
  multi\_classif\_fht & 0.394 (0.00586) & 0.359 (0.00462) & \textbf{0.321} (0.00597) & 0.373 (0.00415) \\ 
   \hline
poisson\_r & 7060 (78.2) & 7000 (75.4) & \textbf{6970} (73.7) & 6990 (68.1) \\ 
  poisson\_f3 & 11800 (45.5) & 11700 (45.1) & \textbf{11700} (50) & 11700 (51.2) \\ 
  gamma\_r & 4610 (159) & 4620 (160) & \textbf{4610} (158) & 4620 (157) \\ 
  gamma\_f3 & 14300 (36.6) & 14300 (35.5) & \textbf{14300} (35.7) & 14300 (35.1) \\ 
  tobit\_r & 4050 (45.7) & 4040 (48.7) & \textbf{4040} (40.5) &  \\ 
  tobit\_f3 & 4540 (45.7) & 4540 (47.5) & \textbf{4530} (47.7) &  \\ 
  msr\_f3 & 3.33 (0.0107) & \textbf{3.31} (0.0121) & 3.31 (0.0107) &  \\ 
  msr\_r & 2.43 (0.0408) & 2.2 (0.0361) & \textbf{2.16} (0.0327) &  \\ 
   \hline
Av. rank & 3.5 & 2.5 & 1.25 & 2.75 \\ 
   \hline
p-val Friedman test & 0.000112 &  &  &  \\ 
  Adj. p-val Wilcoxon test & 0.00146 & 0.00488 &  & 0.0156 \\ 
   \hline
\hline
\end{tabular}
\endgroup
\caption{Results for simulated data: Average test error rates for classification and test negative log-likelihoods for regression. 
    In parentheses are approximate standard deviations. 
    Below are average ranks of the methods over the different datasets (only considering datasets for which all four methods are run). 
    Further, a p-value of a Friedman test with an Iman and Davenport correction for comparing the different algorithms is reported.
    The last row shows Holm-Bonferroni corrected p-values of Wilcoxon signed-rank tests for pairwise 
    comparison of Newton boosting with the novel number of weighted samples parameter and the three alternative methods.} 
\label{results_simulateddata}
\end{table}

The results are reported in Figure \ref{sim_data_score} and in Table \ref{results_simulateddata}. See Section \ref{realdata} for more details on the plot and table. We find again that Newton boosting has the highest predictive accuracy for the large majority of datasets, followed by hybrid gradient-Newton, with gradient boosting having the lowest predictive accuracy. The Friedman test with an Iman and Davenport correction shows that there are statistically significant differences among the different boosting approaches. Further, Newton boosting performs significantly better in terms of predictive accuracy than both gradient and hybrid gradient-Newton boosting despite the relatively small sample size and the multiple testing correction. Finally, Newton boosting with the new equivalent sample size per leaf tuning parameter has higher predictive accuracy than the \texttt{XGBoost} implementation with the unnormalized number of weighted samples per leaf parameter.

\section{Discussion}

\subsection{Does Newton boosting show higher predictive accuracy than gradient boosting due to faster convergence?}
\begin{figure*}[ht!]
	\centering
	\includegraphics[width=0.3\textwidth]{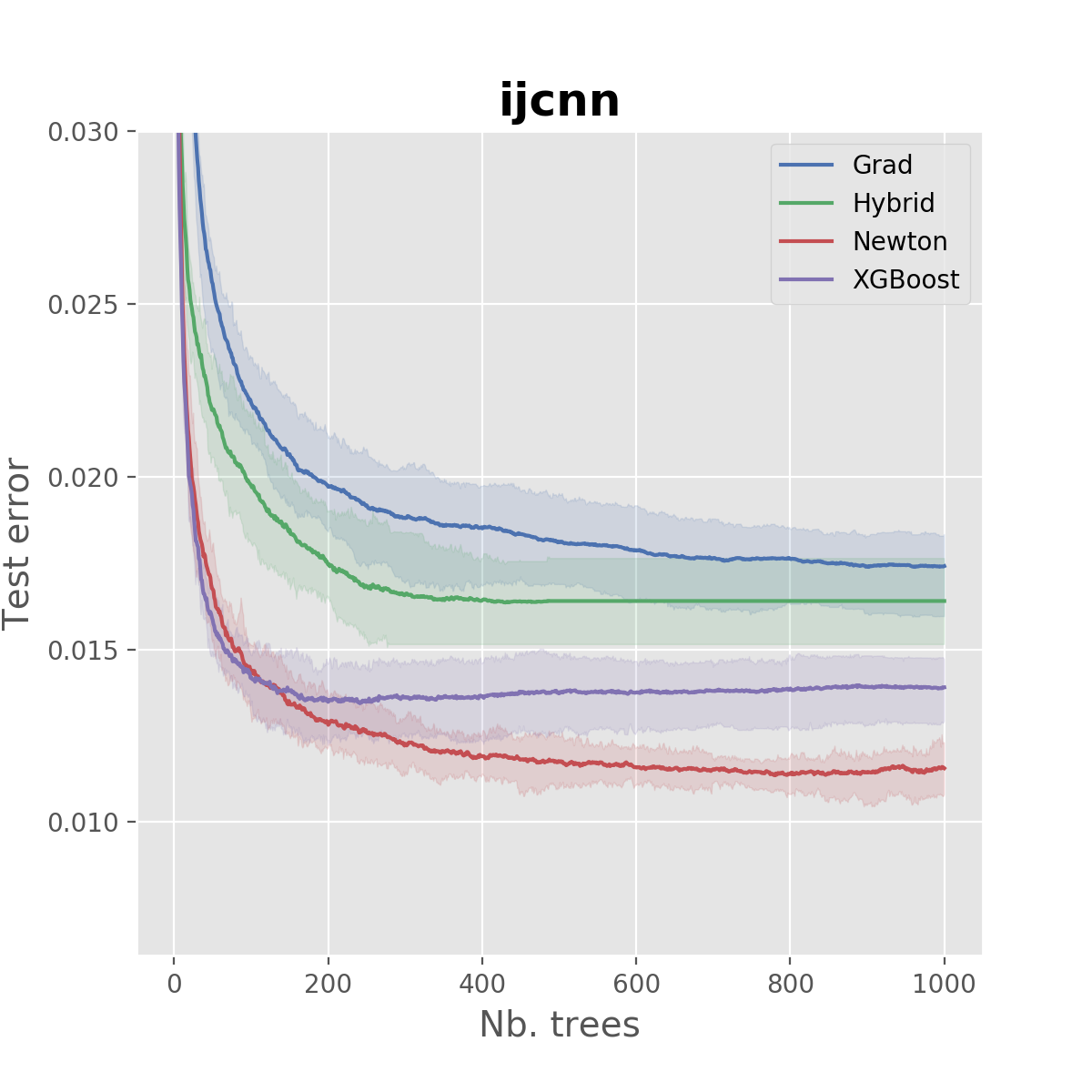}
	\includegraphics[width=0.3\textwidth]{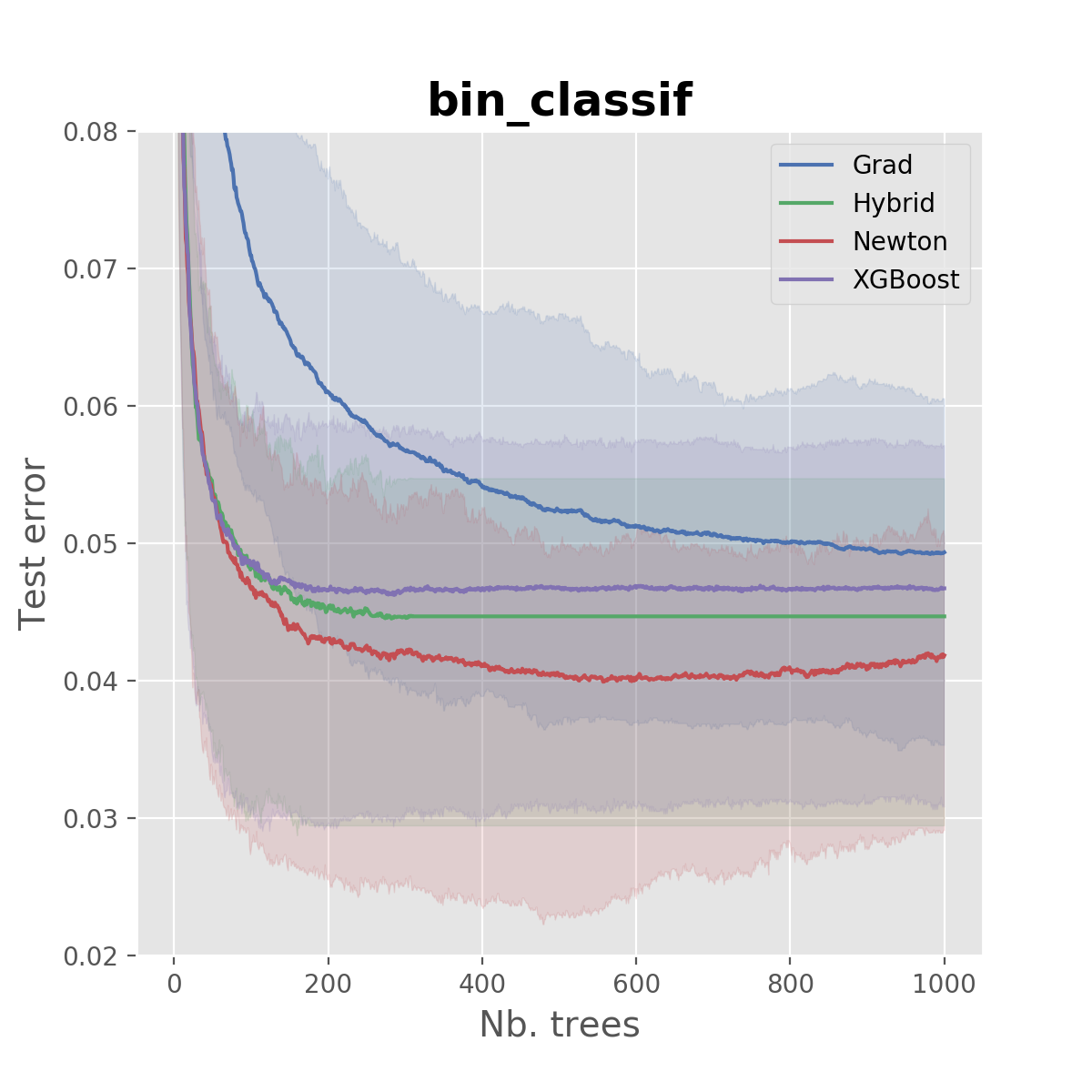}
	\includegraphics[width=0.3\textwidth]{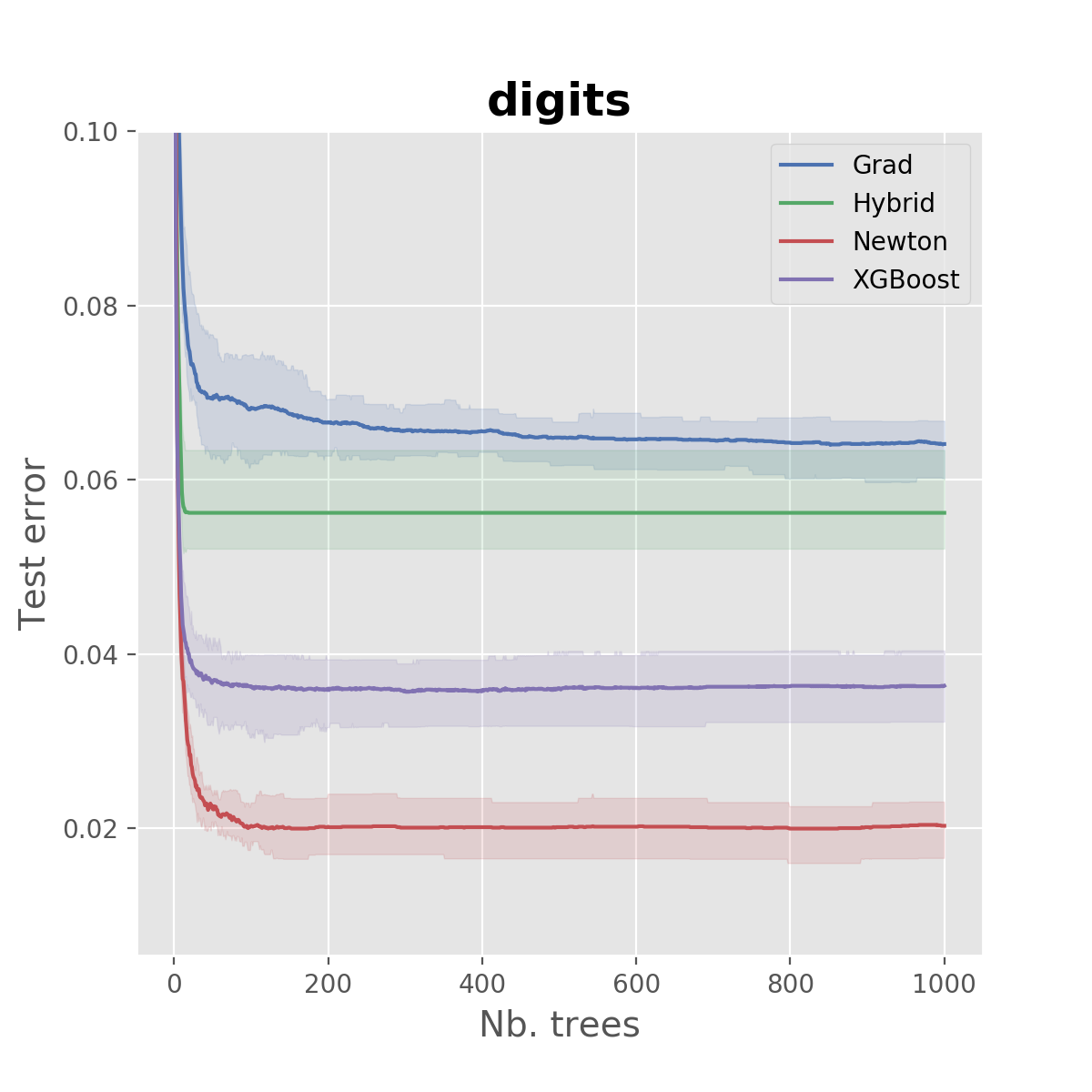}
	\includegraphics[width=0.3\textwidth]{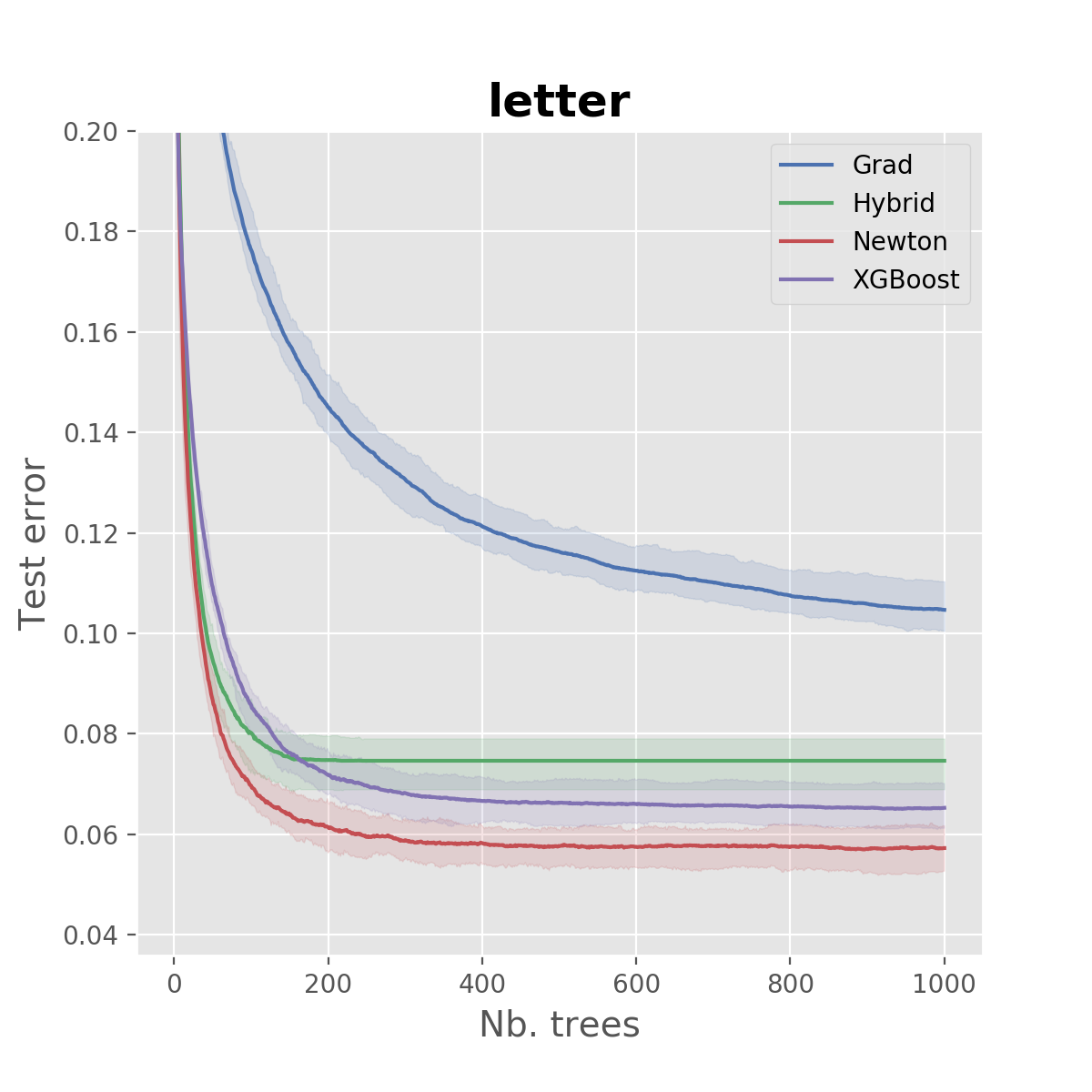}
	\includegraphics[width=0.3\textwidth]{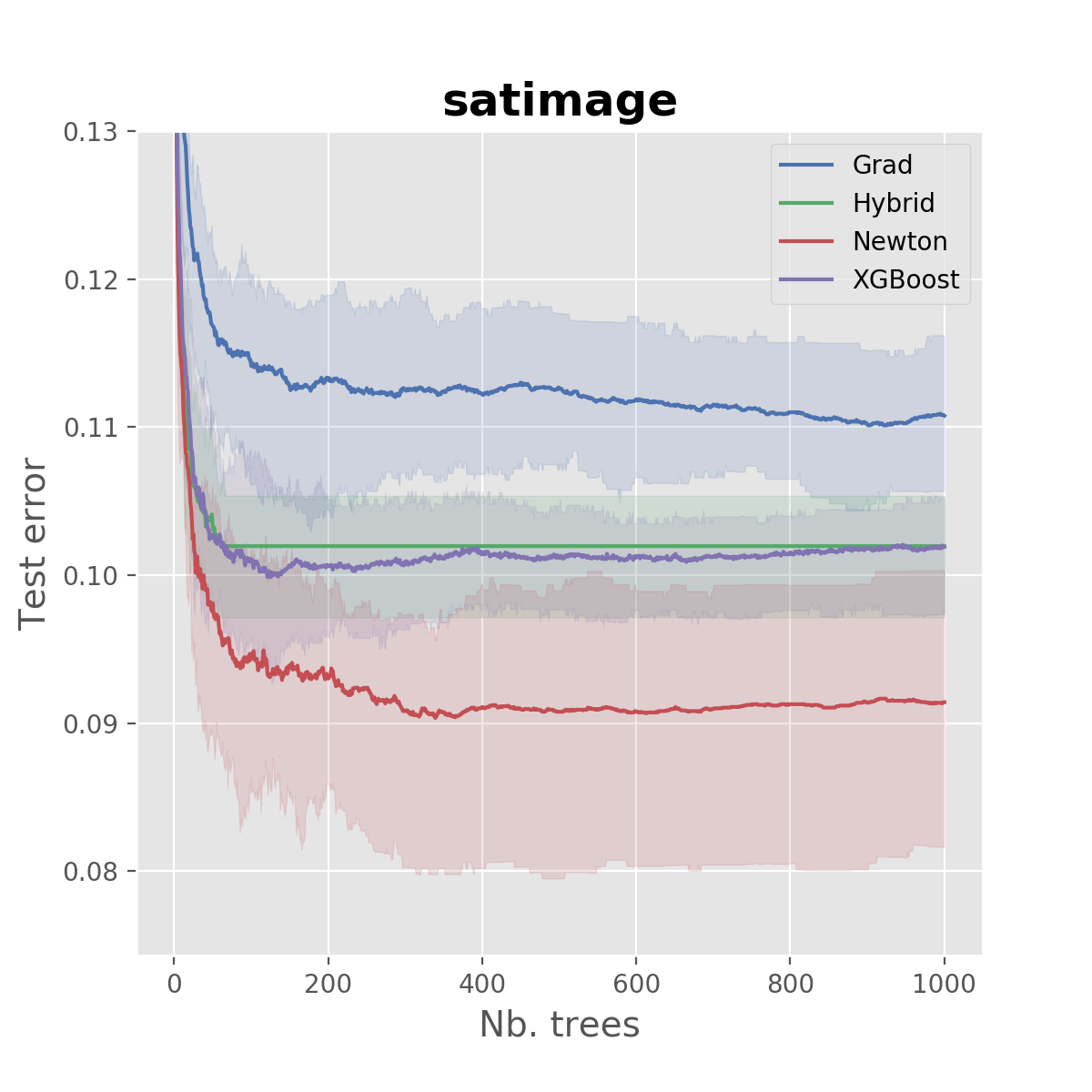}
	\includegraphics[width=0.3\textwidth]{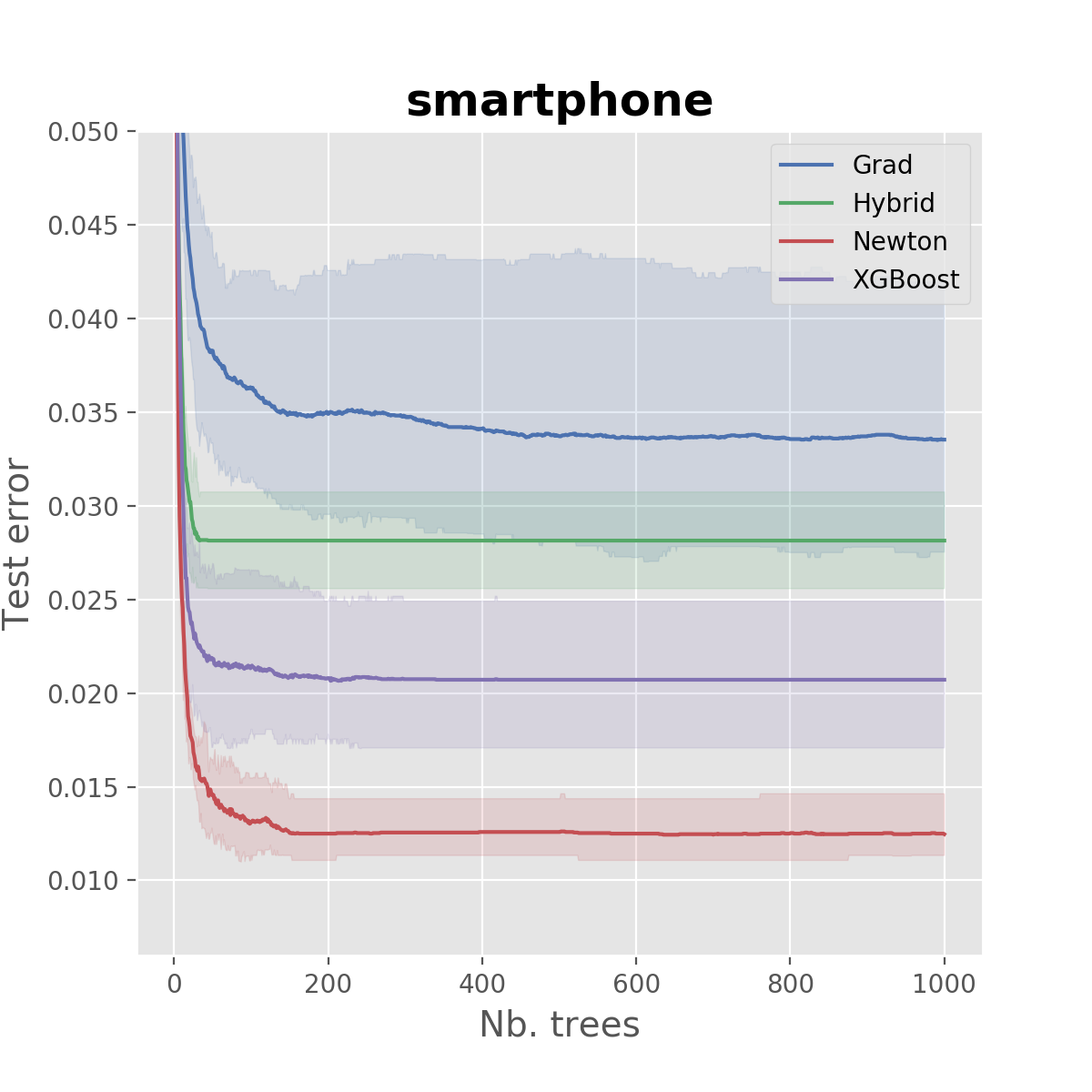}
	\includegraphics[width=0.3\textwidth]{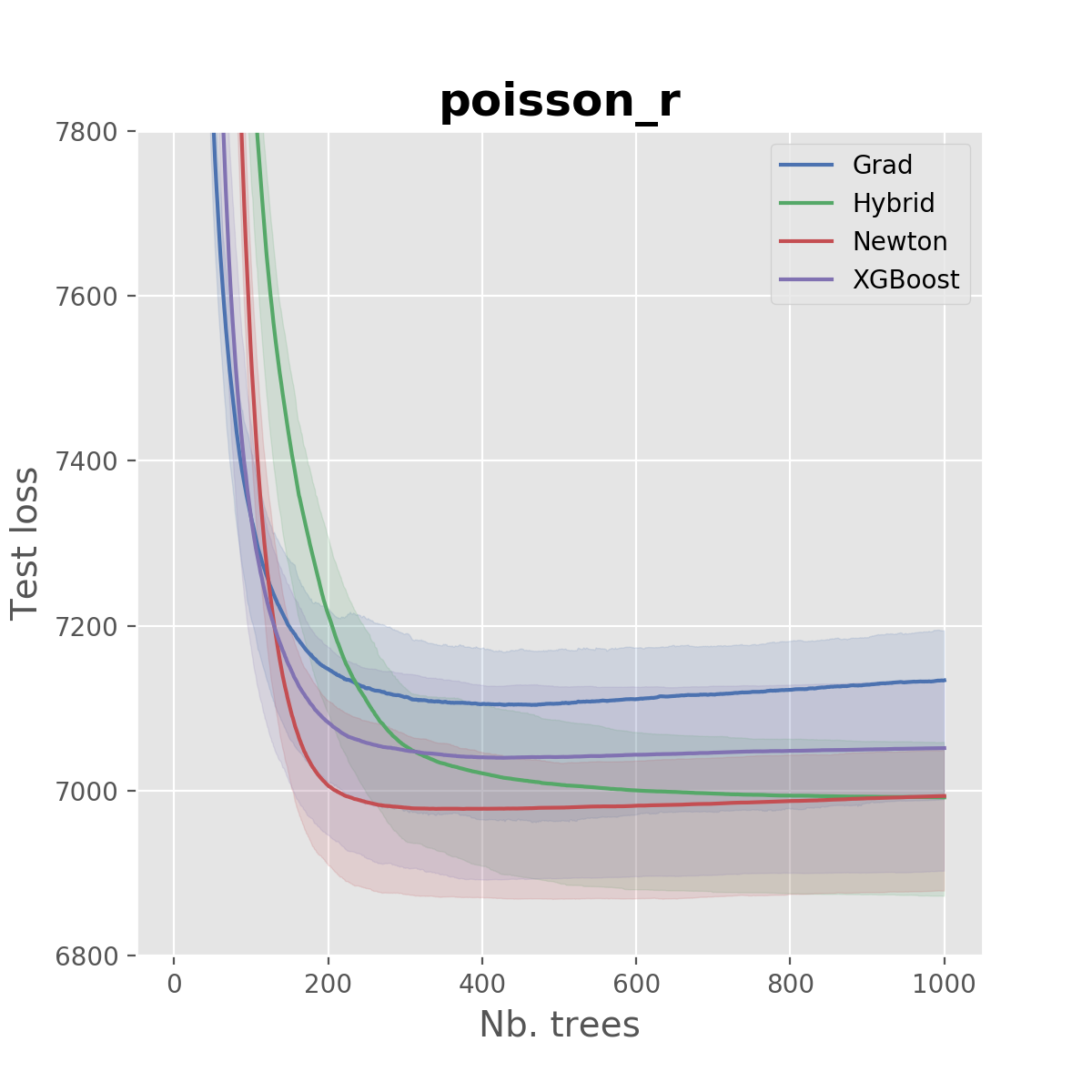}
	\includegraphics[width=0.3\textwidth]{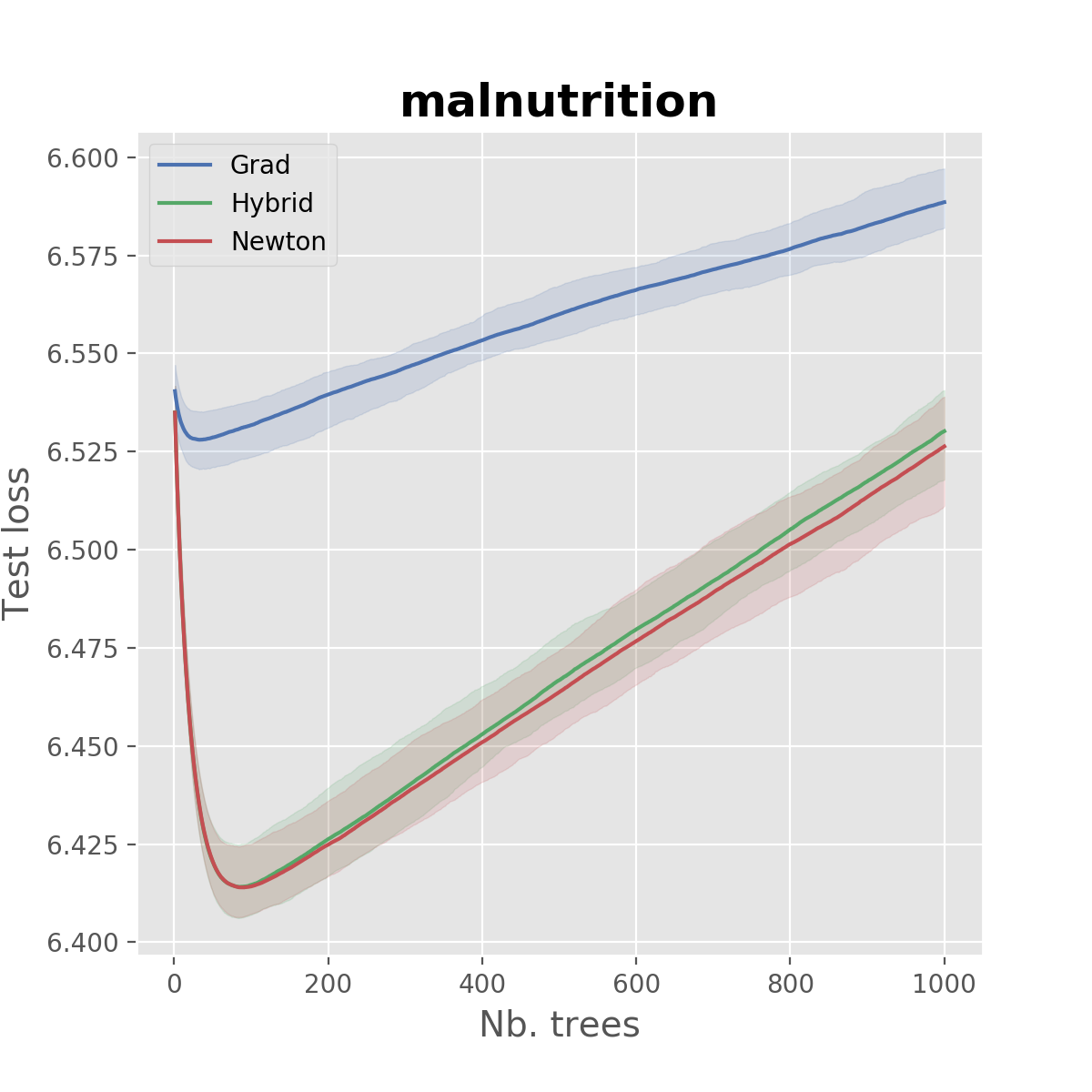}
	\includegraphics[width=0.3\textwidth]{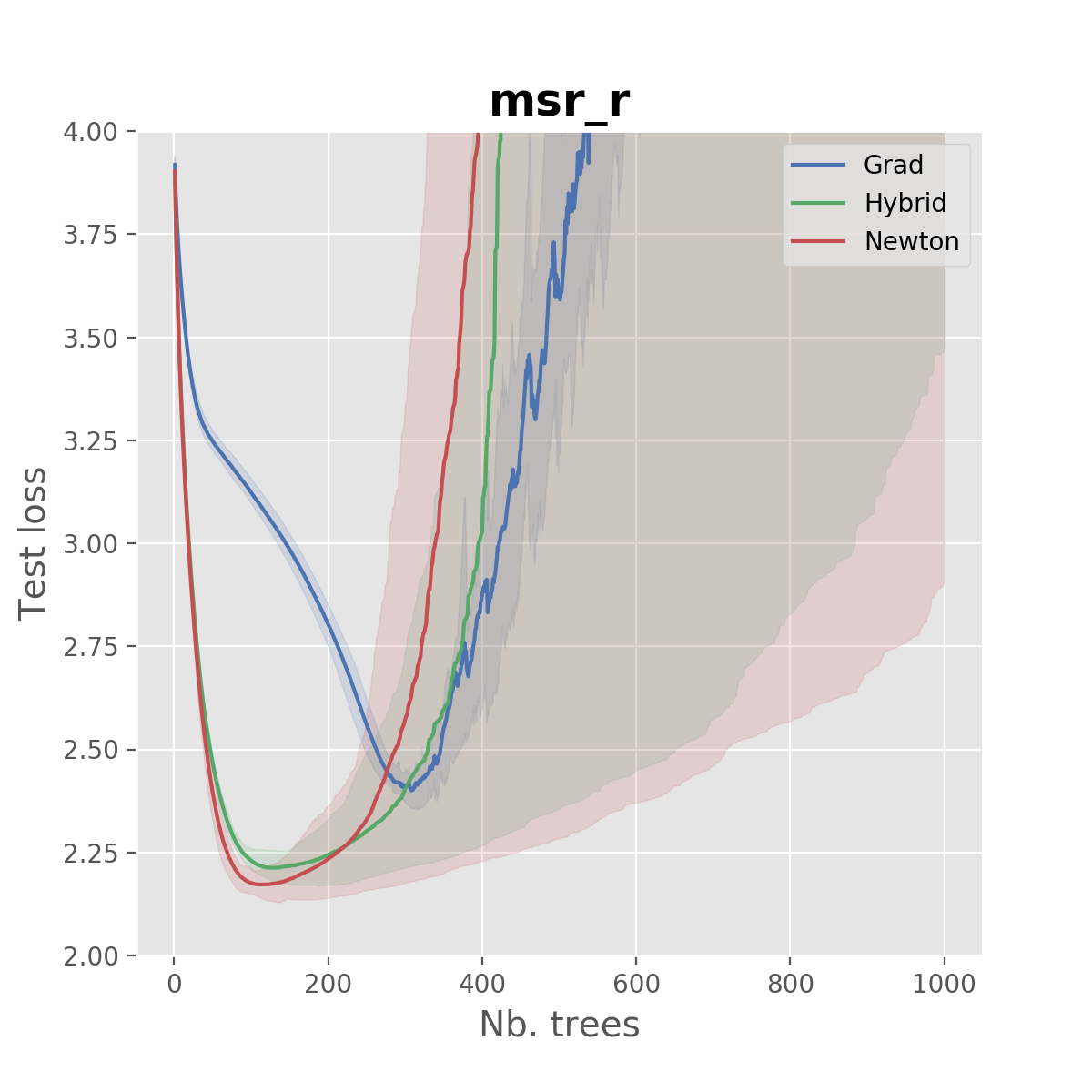}
	\caption{Test error rate (classification) and test negative log-likelihood (regression) versus boosting iteration number.}
	\label{trace_plot} 
\end{figure*}

\begin{figure*}[ht!]
	\centering
	\includegraphics[width=0.3\textwidth]{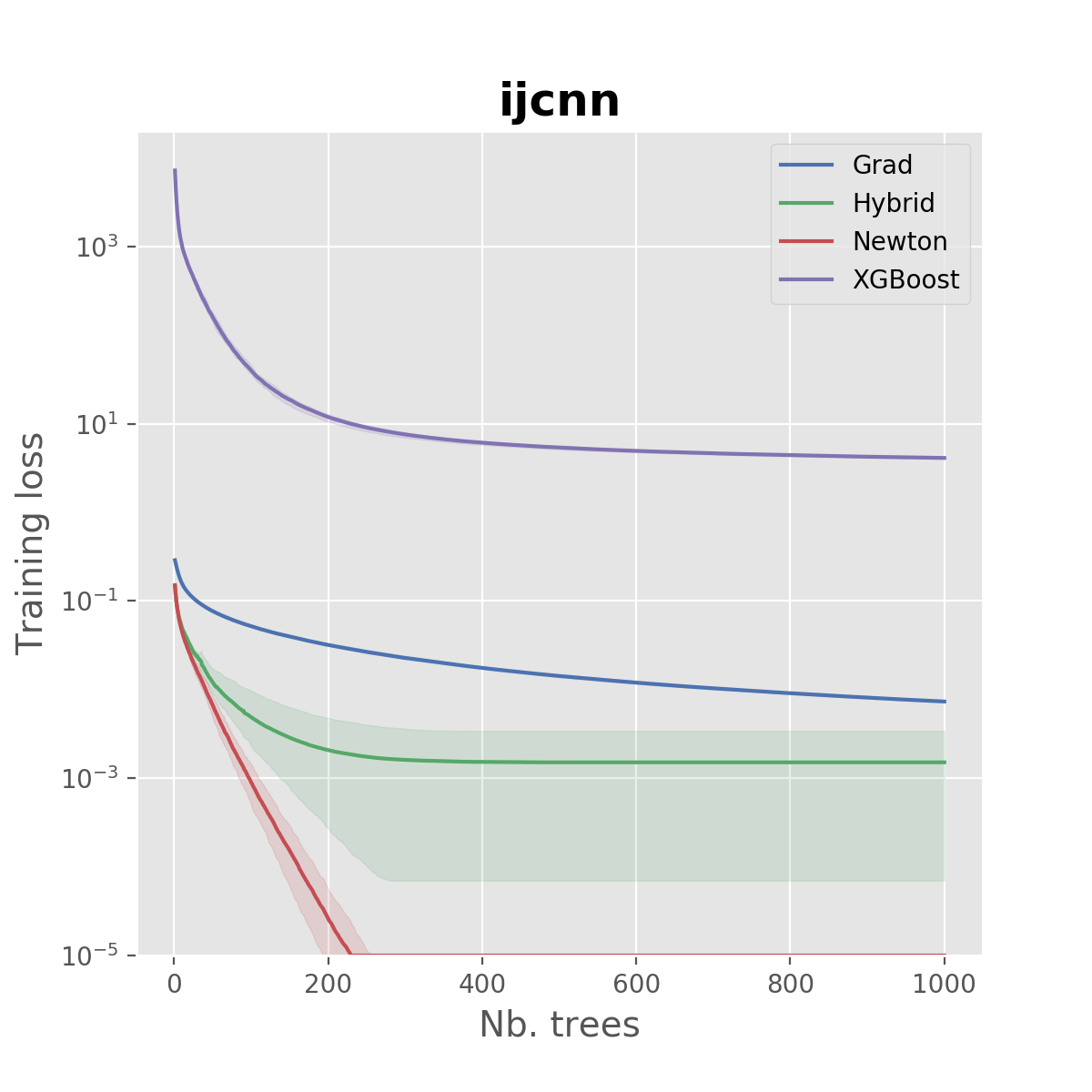}
	\includegraphics[width=0.3\textwidth]{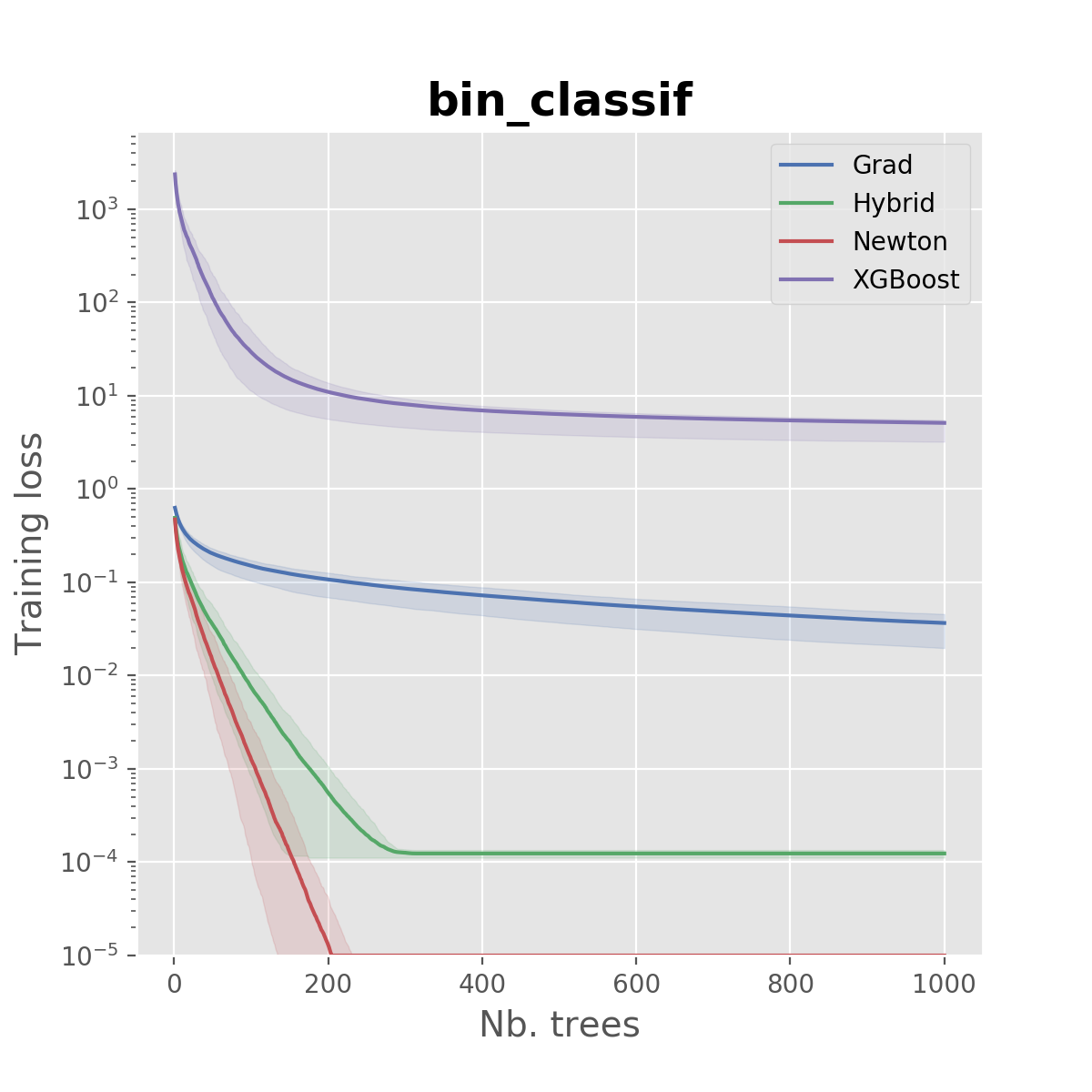}
	\includegraphics[width=0.3\textwidth]{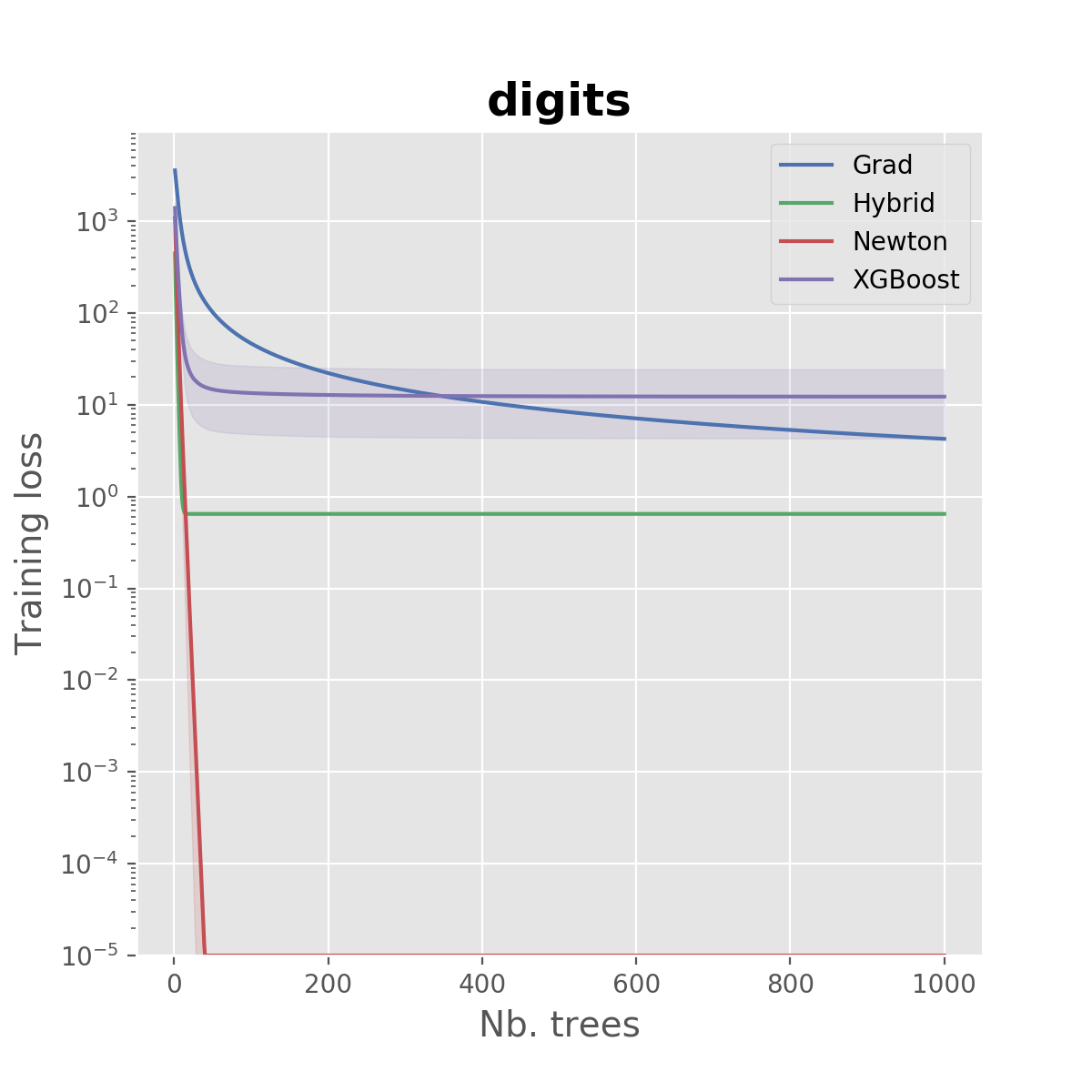}
	\includegraphics[width=0.3\textwidth]{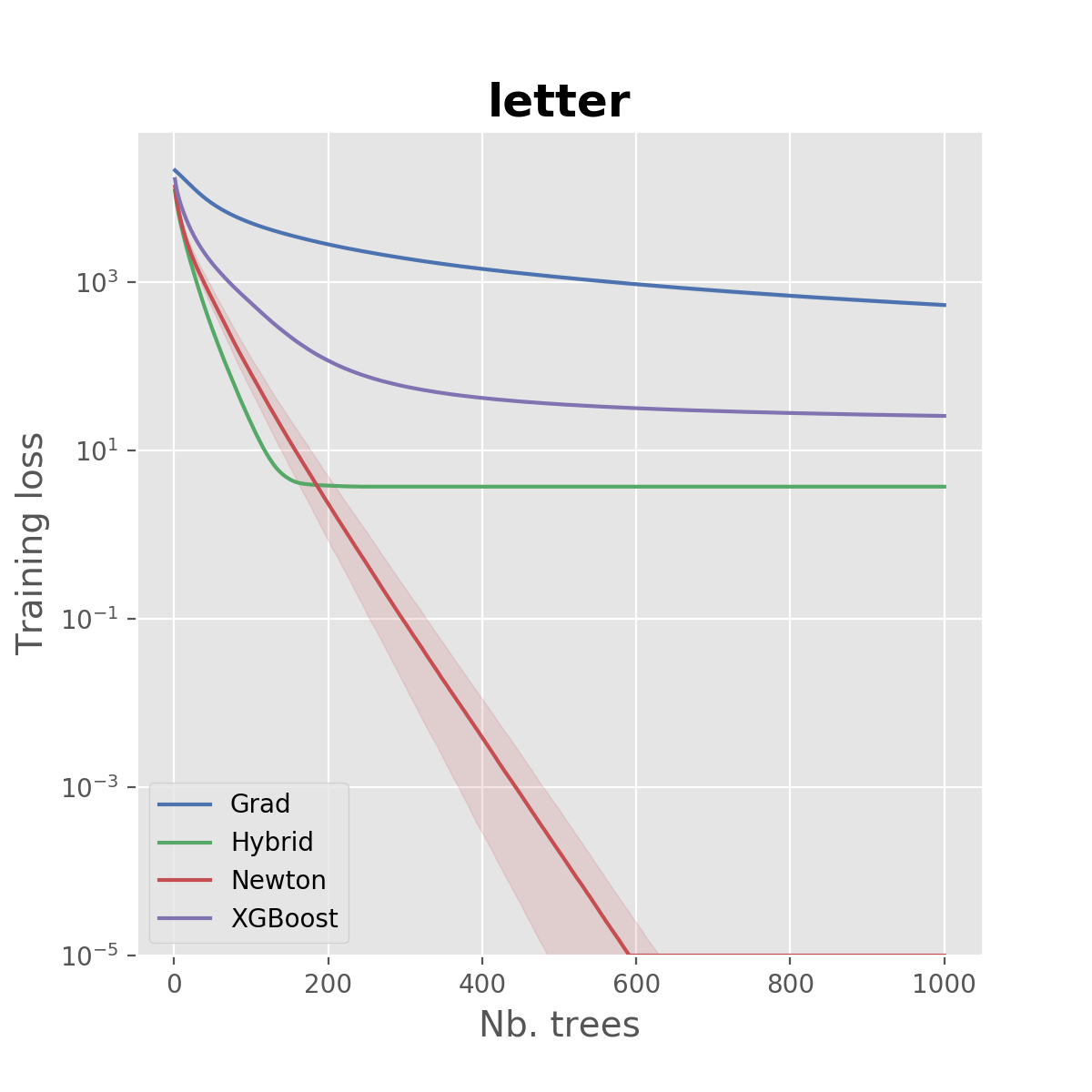}
	\includegraphics[width=0.3\textwidth]{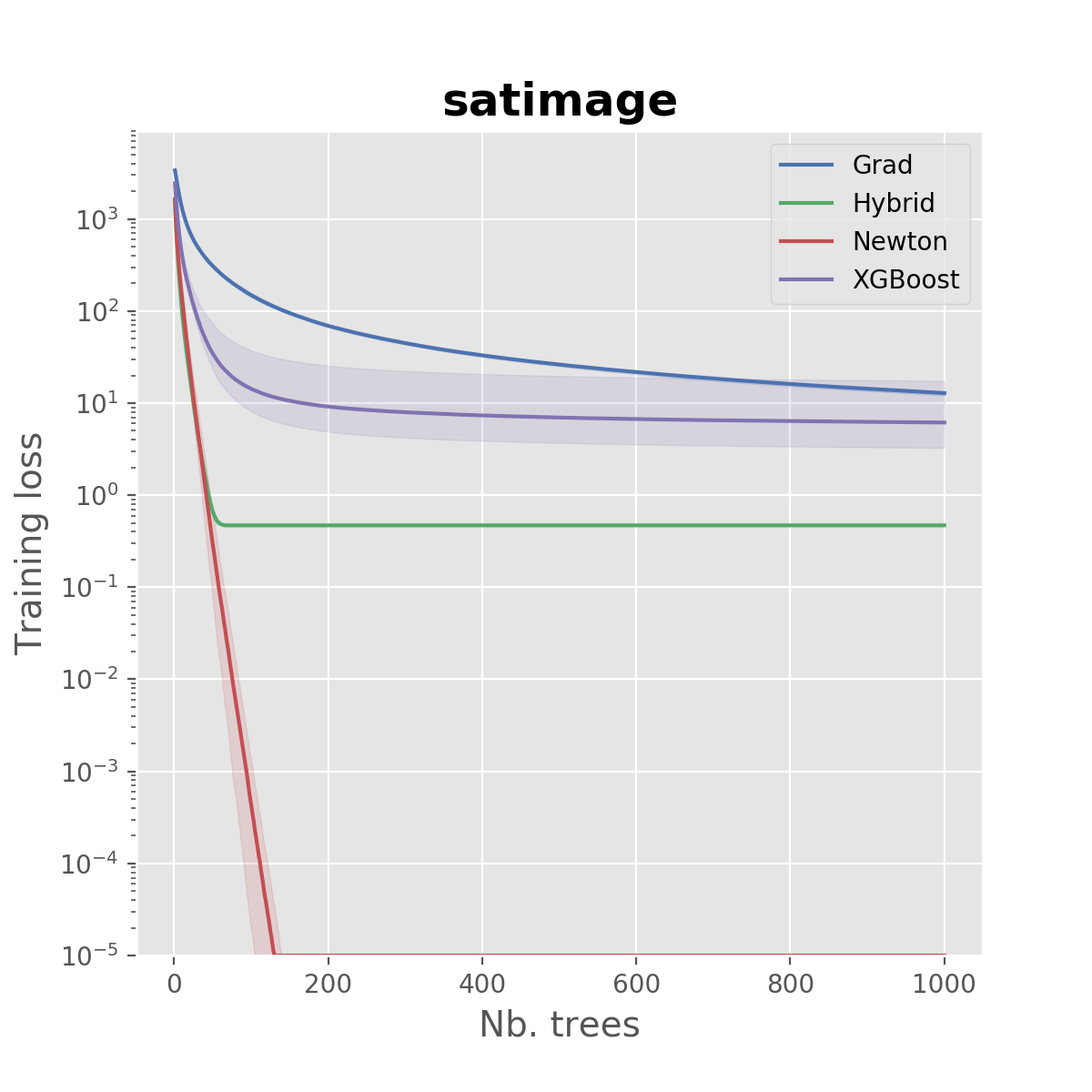}
	\includegraphics[width=0.3\textwidth]{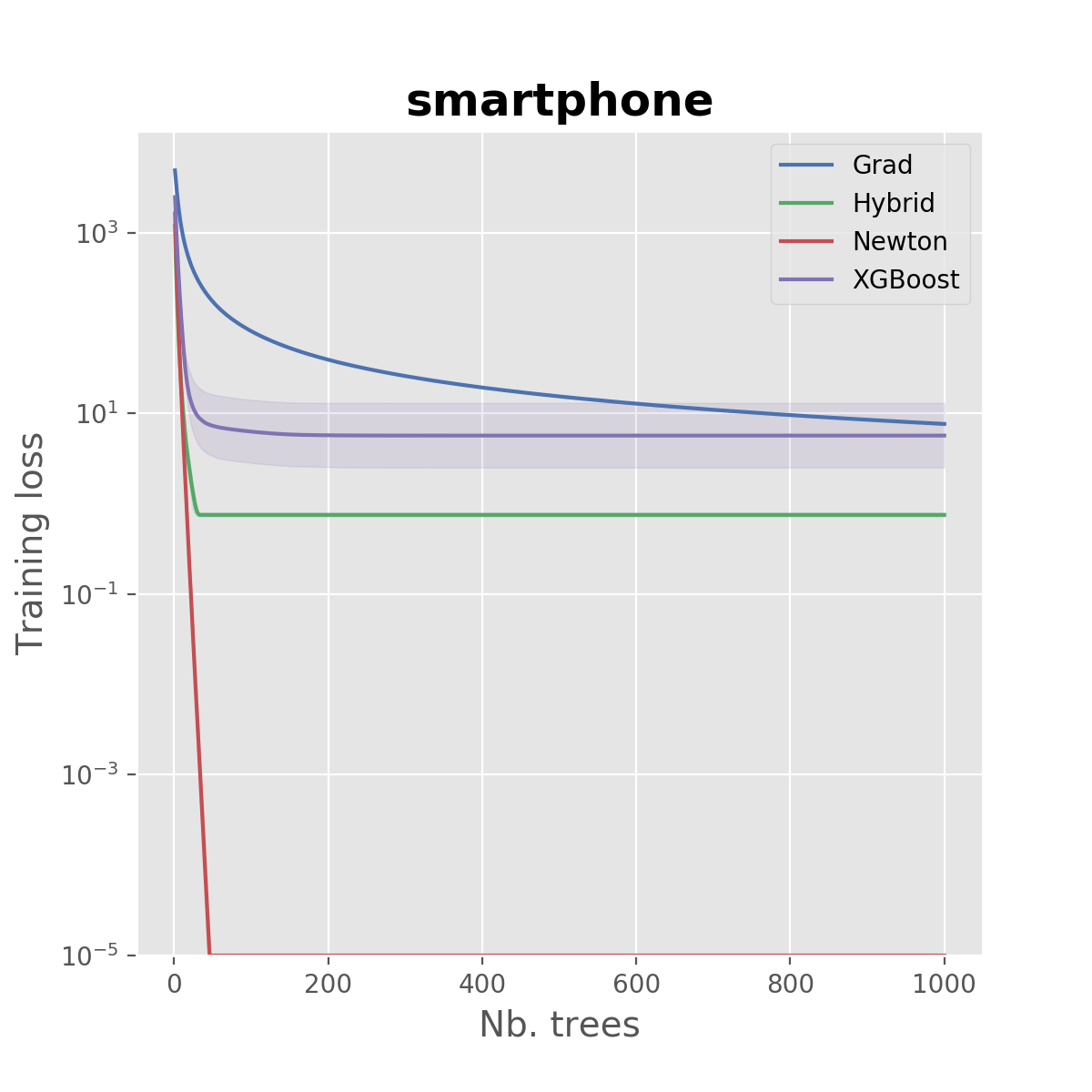}
	\includegraphics[width=0.3\textwidth]{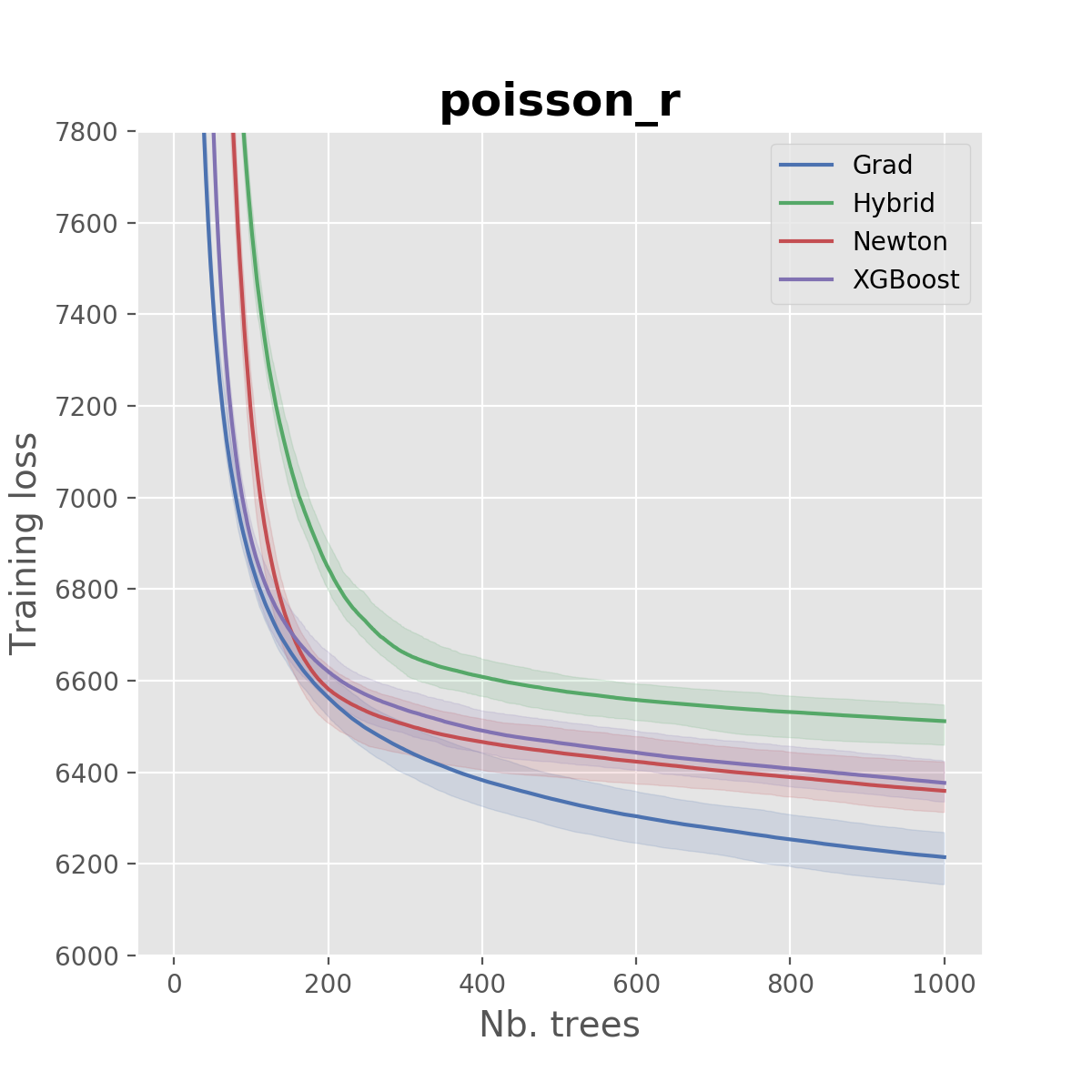}
	\includegraphics[width=0.3\textwidth]{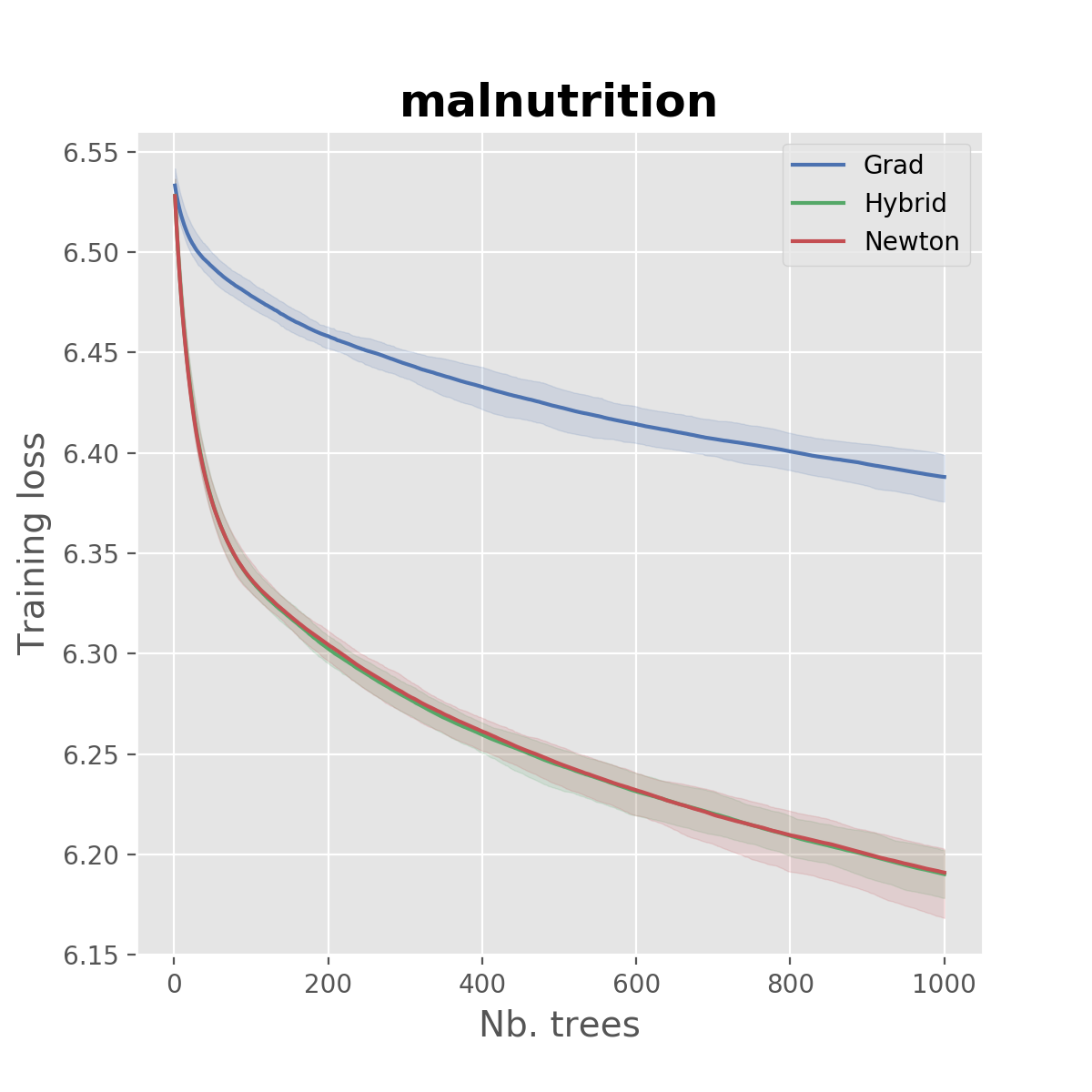}
	\includegraphics[width=0.3\textwidth]{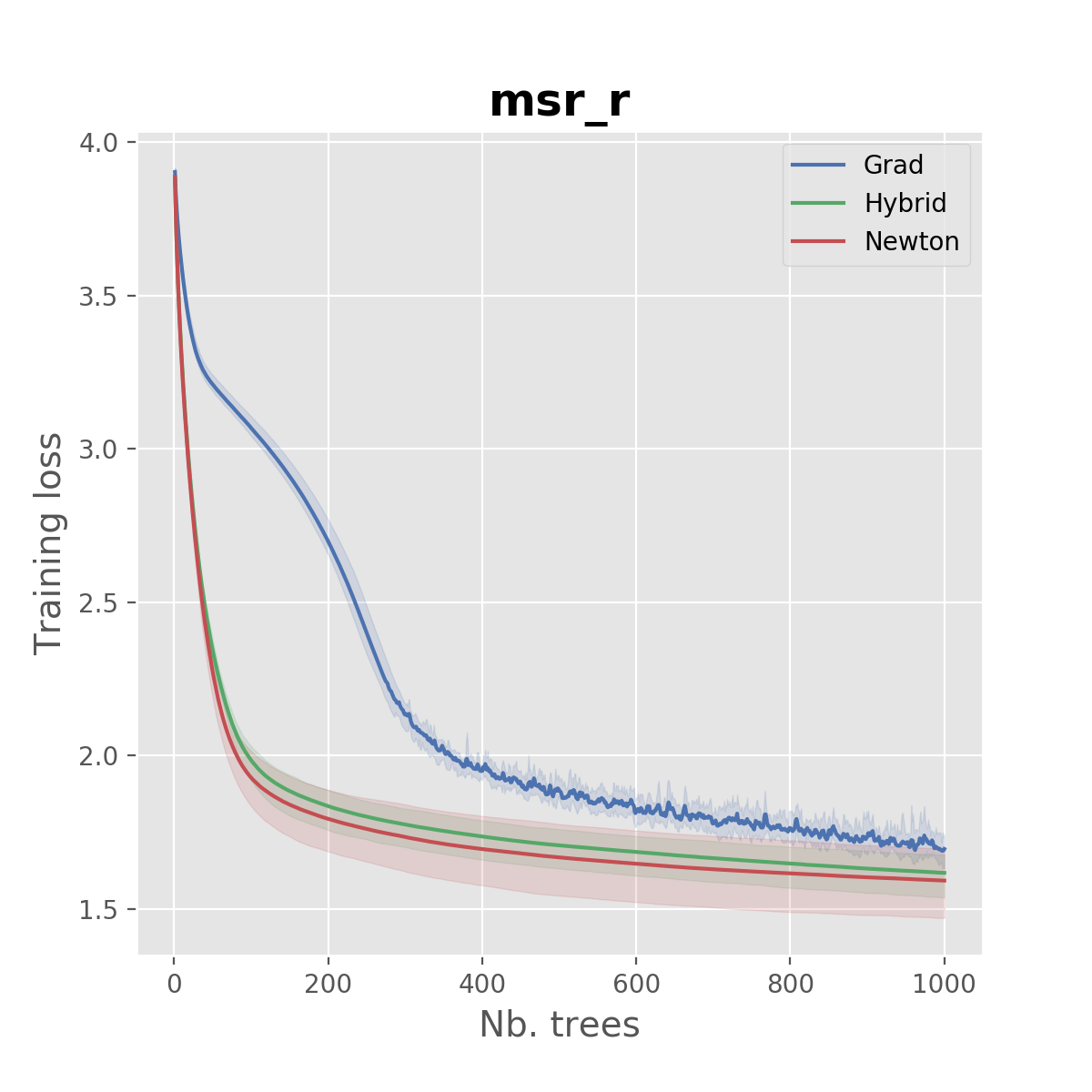}
	\caption{Training loss versus boosting iteration number. The (logarithmic) y-scale is truncated at $10^{-5}$ for better visualization.}
	\label{trace_plot2} 
\end{figure*}

In the previous sections, we have empirically shown that Newton boosting often results in higher predictive accuracy than gradient and also hybrid gradient-Newton boosting. A potential explanation for the observed phenomenon is that Newton boosting converges faster than both gradient and hybrid gradient-Newton boosting, and that hybrid boosting also converges faster than gradient boosting. This, in turn, could allow for using a smaller shrinkage parameter $\nu$, and smaller shrinkage parameters usually lead to increased predictive accuracy. To investigate whether this is the main reason for the differences, we show in Figures \ref{trace_plot} and \ref{trace_plot2} test error rates (classification) and test negative log-likelihoods (regression) as well as training losses versus iteration numbers for several datasets for which we have observed large differences. In order that the results for the different sample splits and also boosting methods are comparable, learning rates are fixed and not tuned. Specifically, we consider the following datasets and learning rates: ijcnn ($\nu=0.5$), bin\_classif ($\nu=0.5$), digits ($\nu=0.5$), letter ($\nu=0.1$), satimage ($\nu=0.3$), smartphone ($\nu=0.5$), poisson\_r ($\nu=0.03$), malnutrition ($\nu=0.03$), and msr\_r ($\nu=0.05$). Note that this list of datasets includes both binary and multiclass classification as well as Poisson and mean-scale regression tasks. The solid lines in Figures \ref{trace_plot2} and \ref{trace_plot} represent means over ten different data splits into equally sized training, validation, and test data. The lower and upper values of the shaded areas are obtained after point-wise discarding the lowest and largest values. Training losses are shown on a logarithmic scale with a lower cap at $10^{-5}$ for better visualization. 

As expected, Figures \ref{trace_plot} and \ref{trace_plot2} show that gradient boosting often converges slower than Newton and also hybrid gradient-Newton boosting. Concerning the latter two, we visually observe almost no difference in convergence speed. However, the plots also show that the faster convergence speed of Newton boosting is likely not the reason for the observed differences in predictive accuracy since the minima of the test errors for Newton boosting are usually achieved after fewer iterations compared to gradient boosting. Interestingly, our results indicate that Newton boosting converges to lower training losses, which are essentially zero for the majority of classification datasets, while at the same time having higher test accuracy. The fact that interpolating classifiers with zero training loss generalize well to novel data seems to be in contradiction to the well-known bias-variance trade-off. However, similar results have recently been observed for other datasets and complex models such as deep learning and kernel machines \citep{zhang2016understanding,belkin18a,belkin18b}. For the Poisson regression dataset, we observe that gradient boosting results in a lower training loss but a higher test loss. For the mean-scale regression datasets, we find that Newton and hybrid boosting show lower test and training losses than gradient boosting, and, in contrast to the classification datasets, we find clear signs of overfitting as the test losses start to increase again after a certain number of boosting iterations. Concerning the \texttt{XGBoost} implementation, we find that \texttt{XGBoost} results in higher training losses and also higher test errors compared to our Newton boosting implementation. In particular, the training losses do not converge to zero for the classification datasets.

Finally, we recall that in cases where the Hessians $h_{m,i}$ defined in Equation \eqref{hessdata} are constant, there is no difference between gradient and Newton boosting. It is thus likely that the more variation there is in the second-order terms $h_{m,i}$ the larger is the difference between gradient and Newton boosting.

\subsection{Importance of the minimum number of (weighted) samples per leaf parameter}\label{notunemain}
In Table \ref{results_real_noWdata} in Appendix \ref{notune}, we additionally report results for the real-world datasets when the minimum number of (weighted) samples per leaf parameter is not tuned by minimizing a validation loss and simply set to a default value. I.e., for gradient and hybrid gradient-Newton boosting, the minimum number of samples per leaf is one, for Newton boosting with our proposed choice in \eqref{minsamp}, we set the minimum equivalent sample size per leaf parameter to one, and for the \texttt{XGBoost} implementation, we set the minimum sum of Hessians to its default value, i.e., also one.\footnote{We exclude the mean-scale regression datasets as there is no obvious default value since the minimum number of samples per leaf needs to be larger than one.} Overall, we find that the difference in predictive accuracy between Newton boosting and gradient as well as hybrid gradient-Newton boosting is more pronounced when the minimum number of samples parameter is not tuned. Interestingly, we find that Newton boosting with the unnormalized sum of Hessians per leaf parameter as implemented in \texttt{XGBoost} and Newton boosting with the novel equivalent number of weighted samples per leaf parameters as implemented in \texttt{KTBoost} perform now almost equally well, and that the results of Newton boosting with the novel equivalent sample size per leaf parameter are worse compared to the ones in Table \ref{results_realdata} when also tuning this parameter. This provides evidence that the number of samples per leaf is an important tuning parameter, and that the unnormalized version of this tuning parameter is difficult to tune.

\subsection{Additional results and robustness check}\label{addres}
In the following, we report additional results to show that our findings are robust to the choice of the maximal tree depth tuning parameters and also to the sample size.

\subsubsection{Maximal tree depth}\label{maxtreedepth}
We additionally consider the following maximal tree depths: $1$ (stumps), $3$, $8$, and $20$. The results in Appendix \ref{treedepth} show that we continue to observe very similar differences among the different boosting versions also for other maximal tree depths. In particular, Newton boosting results in the highest predictive accuracy. For the majority of the datasets, stumps and also trees with maximal depth $3$ perform worse than larger trees. However, for one simulated dataset (multi\_classif\_fht), stumps result in clearly increased predictive accuracy. Further, very large trees with a depth of $20$ generally perform slightly worse than trees with a depth of $5$ or $8$.

\subsubsection{Simulated data with a smaller sample size}
We repeat the simulation study using a smaller sample size. Specifically, we use $n=500$ samples and do $100$ simulation runs with the same specifications as in Section \ref{simdata}. The results for this are reported in Table \ref{results_simulated_smalldata} in Appendix \ref{smalldata}. Qualitatively, we find similar results as for the larger sample size. When excluding the Tobit and mean-scale regression datasets for which \texttt{XGBoost} cannot be run, Newton boosting has an average rank of $1.25$, hybrid gradient-Newton boosting has an average rank of $2.38$, gradient boosting has an average rank of $3.57$, and \texttt{XGBoost} has an average rank of $2.88$. Further, gradient boosting has significantly lower predictive accuracy than Newton boosting. The differences between hybrid gradient-Newton boosting and Newton boosting are less pronounced and not significant, though. Further, the differences between Newton boosting with the novel equivalent sample size per leaf parameter and the \texttt{XGBoost} implementation of Newton boosting with the unnormalized number of weights parameter are also not significant. We note, however, that these statistical tests are done using very small sample sizes of only $12$ or $8$, respectively, and thus have low power. 

\section{Conclusions}
We compare gradient and Newton boosting as well as a hybrid variant of the two with trees as base learners on a wide range of classification and regression datasets. Our empirical results show that Newton boosting outperforms gradient and often also hybrid gradient-Newton boosting in terms of predictive accuracy. Further, we present empirical evidence that this outperformance is not a consequence of a faster convergence speed of Newton boosting. Interestingly, Newton boosting converges to lower values of the empirical risk while at the same time having lower test errors. In addition, we introduce a novel tuning parameter denoted as \textit{equivalent sample size per leaf} parameter which is interpretable, intuitive to tune, and important for predictive accuracy.

We do not have a full explanation for why Newton boosting shows lower generalization errors, and future research should shed light on the reasons for this. Theoretical results such as bounds on the generalization error could help to find an explanation. Further, future research should investigate whether similar results are found for other types of base learners such as splines \citep{buhlmann2003boosting,hothorn2010model}, when combining different base learners such as trees and kernel machines \citep{sigrist2019KTBoost}, when using a linear model as baseline and adding more flexible base learners in a boosting framework \citep{costa2019failure}, or when combining boosting with Gaussian process or mixed effects models \citep{sigrist2020gaussian}. \ifESWA We also note that we have provided evidence that the outperformance of Newton boosting is not related to the sample size. However, we have not considered very large sample sizes. Future research should develop methodology that allows for applying gradient, Newton, and hybrid gradient-Newton boosting to very large datasets and compare the different boosting versions on very large datasets.

Finally, there exist several methodological articles on boosting algorithms in the expert and intelligent systems literature  \cite[e.g.][]{de2017data,kadkhodaei2020hboost,barsacchi2020analysis}. For instance, \citet{de2017data} consider gradient boosting with linear base learners. It would be interesting to analyze how Newton boosting performs in their setting. \citet{kadkhodaei2020hboost} propose a heterogeneous Boosting-based ensemble classifier (HBoost). They use AdaBoost, which corresponds to Newton boosting with an exponential loss function \citep{friedman2000additive}, to train several ensembles with different base learners and then combine the different classifiers. Further, \citet{barsacchi2020analysis} propose to use fuzzy weak learners in a version of the AdaBoost algorithm. Concerning both of these approaches, it would be interesting to investigate how other loss functions and boosting algorithms perform.

\fi

\section*{Acknowledgments}
This research was partially supported by the Swiss Innovation Agency - Innosuisse (25746.1 PFES-ES). We are grateful to Christoph Hirnschall and Torsten Hothorn for valuable suggestions and discussions.

\bibliography{bibfile_boosting}

\clearpage

\begin{appendices}
	\section{Loss functions for regression and classification tasks}\label{lossfcts}
	In the following, we list the loss functions and corresponding gradients and second derivatives that we consider in this article.
	
	\begin{itemize}
		\item \textbf{Binary classification}\\ 
		$Y|X\sim \text{Bernoulli}(p), ~~p=\left(1+e^{-F(X)}\right)^{-1}$\\
		\textit{Loss}:    $L(Y,F)=-YF+\log\left(1+e^F\right)$\\
		\textit{Gradient}:    $\frac{\partial L(Y,F)}{\partial F}=-Y+p$\\
		\textit{Hessian}:    $\frac{\partial^2L(Y,F)}{\partial F^2}=p(1-p)$
		\item \textbf{Multiclass classification}\\ 
		$Y|X\sim \text{Multinom}(p_1,\dots,p_K), ~~p_k=\frac{e^{F_k(X)}}{\sum_{l=1}^Ke^{F_k(X)}}$, $k=1,\dots,K$\\
		\textit{Loss}:\\    $L(Y,F)=\sum_{k=1}^K\left(-\mathbbm{1}_{\{Y=k\}}F_k+\log\left(\sum_{l=1}^Ke^{F_l(X)}\right)\right)$, $F=(F_1\dots,F_K)$\\
		\textit{Gradient}:    $\frac{\partial L(Y,F)}{\partial F_k}=-\mathbbm{1}_{\{Y=k\}}+p_k$\\
		\textit{Hessian}:    $\frac{\partial^2 L(Y,F)}{\partial F_k^2}=p_k(1-p_k)$\\
		As in \citet{friedman2000additive}, we use $\frac{\partial^2 L(Y,F)}{\partial F_k \partial F_j}=0$ for simplicity.
		\item \textbf{Poisson regression}\\ 
		$Y|X\sim \text{Pois}(\lambda), ~~\lambda=e^{F(X)}$\\
		\textit{Loss}:    $L(Y,F)=-YF+e^F$\\
		\textit{Gradient}:    $\frac{\partial L(Y,F)}{\partial F}=-Y+e^{F}$\\
		\textit{Hessian}:    $\frac{\partial^2L(Y,F)}{\partial F^2}=e^{F}$
		\item \textbf{Gamma regression}\\ 
		$Y|X\sim \text{Gamma}(\gamma,\lambda)$ with shape $\gamma$ and rate $\lambda$, $\lambda=\gamma e^{-F(X)}$\\
		\textit{Loss}:    $L(Y,F)=\gamma\left(F+e^{-F}Y\right)-(\gamma-1)\log(Y)-\gamma\log(\gamma)+\log(\Gamma(\gamma))$\\
		\textit{Gradient}:    $\frac{\partial L(Y,F)}{\partial F}=\gamma\left(1-e^{-F}Y\right)$\\
		\textit{Hessian}:    $\frac{\partial^2L(Y,F)}{\partial F^2}=\gamma e^{-F}Y$
		\item \textbf{Tobit model}\\ 
		$Y|X\sim \text{Tobit}_{\{y_l,y_u\}}(\mu,\sigma^2)$, with mean $\mu$, $\mu=F(X)$, and variance $\sigma^2$ of the latent variable and lower and upper censoring thresholds $y_l$ and $y_u$\\
		\textit{Loss}:    
		\begin{equation*}\small
		\begin{split}L(Y,F)=&-\log\left(\Phi\left(\frac{y_l-F}{\sigma}\right)\right)\mathbbm{1}_{y_l}(Y)\\ &+\left(\frac{(Y-F)^2}{2\sigma^2}+\log(\sigma)+0.5\log(2\pi)\right)\mathbbm{1}_{\{y_l<Y<y_u\}}\\
		&-\log\left(1-\Phi\left(\frac{y_u-F}{\sigma}\right)\right)\mathbbm{1}_{y_u}(Y)
		\end{split}
		\end{equation*}
		\textit{Gradient}:    
		\begin{equation*}\small
		\begin{split}
		\frac{\partial L(Y,F)}{\partial F}=&\frac{\phi\left(\frac{y_l-F}{\sigma}\right)}{\sigma\Phi\left(\frac{y_l-F}{\sigma}\right)}\mathbbm{1}_{y_l}(Y)-\frac{Y-F}{\sigma^2}\cdot\mathbbm{1}_{\{y_l<Y<y_u\}}\\
		&-\frac{\phi\left(\frac{y_u-F}{\sigma}\right)}{\sigma\left(1-\Phi\left(\frac{y_u-F}{\sigma}\right)\right)}\mathbbm{1}_{y_u}(Y)
		\end{split}
		\end{equation*}
		\textit{Hessian}: 
		\begin{equation*}\tiny
		\begin{split}&\frac{\partial^2L(Y,F)}{\partial F^2}\\&=\frac{\phi\left(\frac{y_l-F}{\sigma}\right)}{\sigma^2\Phi^2\left(\frac{y_l-F}{\sigma}\right)}\left(\frac{y_l-F}{\sigma}\Phi\left(\frac{y_l-F}{\sigma}\right)+\phi\left(\frac{y_l-F}{\sigma}\right)\right)\mathbbm{1}_{y_l}(Y)\\ & +\frac{1}{\sigma^2}\mathbbm{1}_{\{y_l<Y<y_u\}}\\
		&-\frac{\phi\left(\frac{y_u-F}{\sigma}\right)}{\sigma^2\left(1-\Phi\left(\frac{y_u-F}{\sigma}\right)\right)^2}\left(\frac{y_u-F}{\sigma}\left(1-\Phi\left(\frac{y_u-F}{\sigma}\right)\right)-\phi\left(\frac{y_u-F}{\sigma}\right)\right)\mathbbm{1}_{y_u}(Y)
		\end{split}
		\end{equation*}
		\item \textbf{Mean-scale regression}\\ 
		$Y|X\sim N(\mu,\sigma^2)$, with mean $\mu=F_1(X)$ and standard deviation $\sigma=e^{F_2(X)}$\\
		\textit{Loss}:    $L(Y,F)=\frac{(Y-F_1)^2}{2e^{2F_2}}+F_2+0.5\log(2\pi)$\\
		\textit{Gradient}:    
		\begin{equation*}
		\begin{split}
		\frac{\partial L(Y,F)}{\partial F_1} &=-\frac{Y-F_1}{e^{2F_2}}\\
		\frac{\partial L(Y,F)}{\partial F_2} &=-\frac{(Y-F_1)^2}{e^{2F_2}}+1
		\end{split}
		\end{equation*}
		\textit{Hessian}:    
		\begin{equation*}
		\begin{split}
		\frac{\partial^2 L(Y,F)}{\partial F_1^2} &=\frac{1}{e^{2F_2}}\\
		\frac{\partial^2 L(Y,F)}{\partial F_2^2} &=2\frac{(Y-F_1)^2}{e^{2F_2}}
		\end{split}
		\end{equation*}
		Similarly as for multiclass classification, we assume for simplicity zero off-diagonals for the Hessian, i.e., $\frac{\partial^2 L(Y,F)}{\partial F_1\partial F_2} =0$.
	\end{itemize}
	
	\clearpage
	
	\section{Using a default value for the minimum number of (weighted) samples per leaf parameter}\label{notune}
	
\begin{table}[ht!]
\centering
\begingroup\footnotesize
\begin{tabular}{lllll}
  \hline
\hline
Data & Grad & Hybrid & Newton & XGBoost \\ 
  \hline
adult & 0.129 (0.00211) & \textbf{0.128} (0.00174) & 0.128 (0.0019) & 0.128 (0.00159) \\ 
  bank & 0.101 (0.00196) & 0.101 (0.00162) & 0.101 (0.00162) & \textbf{0.1} (0.00208) \\ 
  cancer & 0.0551 (0.015) & 0.0504 (0.0163) & 0.0452 (0.0136) & \textbf{0.0381} (0.0117) \\ 
  ijcnn & 0.0172 (0.00105) & 0.0148 (0.000824) & \textbf{0.0128} (0.000577) & 0.013 (0.000838) \\ 
  ionosphere & 0.121 (0.0325) & 0.118 (0.036) & 0.107 (0.0357) & \textbf{0.102} (0.029) \\ 
  sonar & 0.307 (0.0637) & 0.308 (0.0595) & 0.289 (0.0514) & \textbf{0.254} (0.0548) \\ 
  car & 0.0567 (0.0103) & 0.0523 (0.0121) & \textbf{0.0407} (0.0122) & 0.0445 (0.0114) \\ 
  covtype & 0.16 (0.00355) & 0.16 (0.00377) & \textbf{0.154} (0.0035) & 0.159 (0.00371) \\ 
  digits & 0.0657 (0.00663) & 0.0489 (0.0051) & \textbf{0.0295} (0.00406) & 0.0362 (0.0035) \\ 
  glass & 0.381 (0.0585) & 0.372 (0.0576) & 0.357 (0.0584) & \textbf{0.348} (0.0582) \\ 
  letter & 0.0917 (0.00558) & 0.075 (0.00424) & \textbf{0.0594} (0.00357) & 0.066 (0.00409) \\ 
  satimage & 0.114 (0.00601) & 0.105 (0.00594) & \textbf{0.0971} (0.00614) & 0.102 (0.00639) \\ 
  smartphone & 0.033 (0.00482) & 0.0238 (0.00318) & \textbf{0.0159} (0.00242) & 0.0197 (0.00265) \\ 
  usps & 0.0782 (0.00265) & 0.0613 (0.00364) & \textbf{0.0435} (0.00271) & 0.054 (0.00341) \\ 
   \hline
insurance & 51900 (308) & 51700 (306) & \textbf{51600} (300) & 51700 (305) \\ 
   \hline
Av. rank & 3.8 & 2.93 & 1.6 & 1.67 \\ 
   \hline
p-val Friedman test & 2.24e-10 &  &  &  \\ 
  Adj. p-val Wilcoxon test & 0.000366 & 0.00061 &  & 0.489 \\ 
   \hline
\hline
\end{tabular}
\endgroup
\caption{Results for real-world data when the minimum number of (weighted)
   samples per leaf parameter is not tuned and set to a default value: Average test error rates for classification and test negative log-likelihoods for regression. 
    In parentheses are approximate standard deviations. 
    Below are average ranks of the methods over the different datasets.
    Further, a p-value of a Friedman test with an Iman and Davenport correction for comparing the different algorithms is reported.
    The last row shows Holm-Bonferroni corrected p-values of Wilcoxon signed-rank tests for pairwise 
    comparison of Newton boosting with the novel number of weighted samples parameter and the three alternative methods.} 
\label{results_real_noWdata}
\end{table}

	\clearpage
	
	\section{Results for different maximal tree depths}\label{treedepth}
	
\begin{table}[ht!]
\centering
\begingroup\footnotesize
\begin{tabular}{lllll}
  \hline
\hline
Data & Grad & Hybrid & Newton & XGBoost \\ 
  \hline
bin\_classif & \textbf{0.15} (0.0466) & 0.151 (0.0467) & 0.151 (0.0483) & 0.151 (0.0476) \\ 
  multi\_classif\_fht & 0.328 (0.00691) & \textbf{0.191} (0.00528) & 0.211 (0.0223) & 0.381 (0.0173) \\ 
  digits & 0.0509 (0.00484) & 0.0426 (0.00563) & \textbf{0.0423} (0.00373) & 0.0582 (0.0155) \\ 
  satimage & 0.126 (0.00587) & 0.12 (0.00624) & \textbf{0.117} (0.00636) & 0.137 (0.0138) \\ 
  ijcnn & 0.0422 (0.00124) & 0.0401 (0.00118) & \textbf{0.04} (0.00127) & 0.0414 (0.00149) \\ 
   \hline
poisson\_r & 6930 (79.2) & 6950 (62.7) & \textbf{6930} (69.7) & 11300 (859) \\ 
  gamma\_r & 4600 (159) & \textbf{4600} (160) & 4600 (161) & 5090 (820) \\ 
  tobit\_r & 4440 (45.7) & 4430 (48.1) & \textbf{4430} (47.9) &  \\ 
  msr\_r & 2.73 (0.0271) & 2.71 (0.0256) & \textbf{2.71} (0.0263) &  \\ 
  malnutrition & 6.53 (0.00615) & 6.41 (0.00739) & \textbf{6.41} (0.00756) &  \\ 
   \hline
Av. rank & 2.57 & 2.07 & 1.79 & 3.57 \\ 
   \hline
p-val Friedman test & 0.196 &  &  &  \\ 
  Adj. p-val Wilcoxon test & 0.168 & 0.343 &  & 0.0938 \\ 
   \hline
\hline
\end{tabular}
\endgroup
\caption{Results when the maximal tree depth is one (stumps): Average test error rates for classification and test negative log-likelihoods for regression. 
    In parentheses are approximate standard deviations. 
    Below are average ranks of the methods over the different datasets (only considering datasets for which all four methods are run). 
    Further, a p-value of a Friedman test with an Iman and Davenport correction for comparing the different algorithms is reported.
    The last row shows Holm-Bonferroni corrected p-values of Wilcoxon signed-rank tests for pairwise 
    comparison of Newton boosting with the novel number of weighted samples parameter and the three alternative methods.} 
\label{results_depth1data}
\end{table}

\begin{table}[ht!]
\centering
\begingroup\footnotesize
\begin{tabular}{lllll}
  \hline
\hline
Data & Grad & Hybrid & Newton & XGBoost \\ 
  \hline
bin\_classif & 0.0558 (0.0132) & 0.0525 (0.0138) & \textbf{0.046} (0.013) & 0.0508 (0.0135) \\ 
  multi\_classif\_fht & 0.38 (0.00335) & 0.344 (0.00968) & \textbf{0.3} (0.00918) & 0.358 (0.00447) \\ 
  digits & 0.0359 (0.00473) & 0.0299 (0.00469) & \textbf{0.024} (0.00342) & 0.0338 (0.00573) \\ 
  satimage & 0.108 (0.00547) & 0.104 (0.00618) & \textbf{0.102} (0.00574) & 0.106 (0.0059) \\ 
  ijcnn & 0.0185 (0.000857) & 0.0167 (0.00116) & \textbf{0.0141} (0.00113) & 0.0158 (0.00083) \\ 
   \hline
poisson\_r & 7030 (60.1) & 6950 (65.3) & \textbf{6940} (64.4) & 6990 (54.3) \\ 
  gamma\_r & \textbf{4600} (159) & 4610 (158) & 4610 (158) & 4640 (156) \\ 
  tobit\_r & 4060 (46) & 4070 (47.9) & \textbf{4060} (50.4) &  \\ 
  msr\_r & 2.42 (0.0281) & 2.36 (0.0236) & \textbf{2.34} (0.0202) &  \\ 
  malnutrition & 6.53 (0.00648) & 6.41 (0.0077) & \textbf{6.41} (0.00798) &  \\ 
   \hline
Av. rank & 3.57 & 2.29 & 1.29 & 2.86 \\ 
   \hline
p-val Friedman test & 0.000886 &  &  &  \\ 
  Adj. p-val Wilcoxon test & 0.129 & 0.129 &  & 0.0469 \\ 
   \hline
\hline
\end{tabular}
\endgroup
\caption{Results when the maximal tree depth is three: Average test error rates for classification and test negative log-likelihoods for regression. 
    In parentheses are approximate standard deviations. 
    Below are average ranks of the methods over the different datasets (only considering datasets for which all four methods are run). 
    Further, a p-value of a Friedman test with an Iman and Davenport correction for comparing the different algorithms is reported.
    The last row shows Holm-Bonferroni corrected p-values of Wilcoxon signed-rank tests for pairwise 
    comparison of Newton boosting with the novel number of weighted samples parameter and the three alternative methods.} 
\label{results_depth3data}
\end{table}

\begin{table}[ht!]
\centering
\begingroup\footnotesize
\begin{tabular}{lllll}
  \hline
\hline
Data & Grad & Hybrid & Newton & XGBoost \\ 
  \hline
bin\_classif & 0.0464 (0.0122) & 0.0408 (0.0104) & \textbf{0.0379} (0.0101) & 0.0434 (0.0111) \\ 
  multi\_classif\_fht & 0.401 (0.00676) & 0.366 (0.00593) & \textbf{0.337} (0.00748) & 0.389 (0.0118) \\ 
  digits & 0.0351 (0.00589) & 0.0297 (0.00465) & \textbf{0.0234} (0.00445) & 0.0384 (0.00427) \\ 
  satimage & 0.105 (0.00579) & 0.0981 (0.00716) & \textbf{0.0958} (0.00604) & 0.102 (0.00613) \\ 
  ijcnn & 0.0143 (0.000823) & 0.0131 (0.000699) & \textbf{0.0116} (0.000586) & 0.0133 (0.000672) \\ 
   \hline
poisson\_r & 7080 (82.5) & 7020 (75.6) & \textbf{6990} (77.1) & 7010 (72) \\ 
  gamma\_r & 4610 (158) & 4620 (156) & \textbf{4610} (158) & 4620 (159) \\ 
  tobit\_r & 4060 (47.8) & 4050 (48.2) & \textbf{4050} (47.4) &  \\ 
  msr\_r & 2.41 (0.0575) & 2.19 (0.0434) & \textbf{2.17} (0.0273) &  \\ 
  malnutrition & 6.53 (0.00658) & 6.42 (0.00823) & \textbf{6.42} (0.00737) &  \\ 
   \hline
Av. rank & 3.57 & 2.43 & 1 & 3 \\ 
   \hline
p-val Friedman test & 2.73e-05 &  &  &  \\ 
  Adj. p-val Wilcoxon test & 0.00586 & 0.00586 &  & 0.0156 \\ 
   \hline
\hline
\end{tabular}
\endgroup
\caption{Results when the maximal tree depth is eight: Average test error rates for classification and test negative log-likelihoods for regression. 
    In parentheses are approximate standard deviations. 
    Below are average ranks of the methods over the different datasets (only considering datasets for which all four methods are run). 
    Further, a p-value of a Friedman test with an Iman and Davenport correction for comparing the different algorithms is reported.
    The last row shows Holm-Bonferroni corrected p-values of Wilcoxon signed-rank tests for pairwise 
    comparison of Newton boosting with the novel number of weighted samples parameter and the three alternative methods.} 
\label{results_depth8data}
\end{table}

\begin{table}[ht!]
\centering
\begingroup\footnotesize
\begin{tabular}{lllll}
  \hline
\hline
Data & Grad & Hybrid & Newton & XGBoost \\ 
  \hline
bin\_classif & 0.0457 (0.0122) & 0.0415 (0.0112) & \textbf{0.0385} (0.00998) & 0.0436 (0.0103) \\ 
  multi\_classif\_fht & 0.41 (0.00486) & 0.388 (0.00579) & \textbf{0.35} (0.00678) & 0.394 (0.0138) \\ 
  digits & 0.0351 (0.00555) & 0.0285 (0.00427) & \textbf{0.0232} (0.0041) & 0.0386 (0.00477) \\ 
  satimage & 0.104 (0.00604) & 0.1 (0.00746) & \textbf{0.0951} (0.00552) & 0.103 (0.00542) \\ 
  ijcnn & 0.0147 (0.000878) & 0.0137 (0.000396) & \textbf{0.0118} (0.00064) & 0.014 (0.000548) \\ 
   \hline
poisson\_r & 7100 (75.6) & \textbf{7030} (77.3) & 7050 (81.2) & 7060 (75.3) \\ 
  gamma\_r & 4610 (158) & 4640 (159) & \textbf{4610} (158) & 4630 (159) \\ 
  tobit\_r & 4070 (47.1) & \textbf{4050} (46.8) & 4060 (46.3) &  \\ 
  msr\_r & 2.42 (0.0457) & 2.19 (0.0434) & \textbf{2.18} (0.0316) &  \\ 
  malnutrition & 6.53 (0.00688) & 6.42 (0.00783) & \textbf{6.42} (0.00743) &  \\ 
   \hline
Av. rank & 3.57 & 2.14 & 1.14 & 3.14 \\ 
   \hline
p-val Friedman test & 0.000298 &  &  &  \\ 
  Adj. p-val Wilcoxon test & 0.00586 & 0.322 &  & 0.0313 \\ 
   \hline
\hline
\end{tabular}
\endgroup
\caption{Results when the maximal tree depth is twenty: Average test error rates for classification and test negative log-likelihoods for regression. 
    In parentheses are approximate standard deviations. 
    Below are average ranks of the methods over the different datasets (only considering datasets for which all four methods are run). 
    Further, a p-value of a Friedman test with an Iman and Davenport correction for comparing the different algorithms is reported.
    The last row shows Holm-Bonferroni corrected p-values of Wilcoxon signed-rank tests for pairwise 
    comparison of Newton boosting with the novel number of weighted samples parameter and the three alternative methods.} 
\label{results_depth20data}
\end{table}

	\clearpage
	
	\section{Results for the simulated data with a smaller sample size}\label{smalldata}
	
\begin{table}[ht!]
\centering
\begingroup\footnotesize
\begin{tabular}{lllll}
  \hline
\hline
Data & Grad & Hybrid & Newton & XGBoost \\ 
  \hline
bin\_classif & 0.111 (0.026) & 0.103 (0.0248) & \textbf{0.0899} (0.0233) & 0.106 (0.0247) \\ 
  bin\_classif\_fht & 0.248 (0.0227) & 0.246 (0.0214) & \textbf{0.237} (0.0216) & 0.242 (0.0244) \\ 
  multi\_classif & 0.356 (0.0386) & 0.346 (0.035) & \textbf{0.329} (0.0349) & 0.346 (0.0374) \\ 
  multi\_classif\_fht & 0.591 (0.0245) & 0.583 (0.0242) & \textbf{0.558} (0.025) & 0.587 (0.0231) \\ 
   \hline
poisson\_r & 778 (40.2) & 769 (40.5) & 773 (37.3) & \textbf{764} (36.9) \\ 
  poisson\_f3 & 1210 (17.5) & 1210 (17) & \textbf{1200} (18.2) & 1220 (20.2) \\ 
  gamma\_r & 471 (43.5) & 476 (42.9) & \textbf{468} (43.3) & 472 (42.9) \\ 
  gamma\_f3 & 1440 (11.3) & 1440 (11.4) & \textbf{1440} (11.5) & 1440 (11.6) \\ 
  tobit\_r & 445 (18.5) & \textbf{442} (18.7) & 443 (18.8) &  \\ 
  tobit\_f3 & 480 (16.1) & \textbf{478} (16) & 478 (15.6) &  \\ 
  msr\_f3 & 3.38 (0.0354) & \textbf{3.37} (0.0381) & 3.38 (0.0419) &  \\ 
  msr\_r & 3.23 (0.199) & 3.19 (0.254) & \textbf{3.07} (0.236) &  \\ 
   \hline
Av. rank & 3.5 & 2.38 & 1.25 & 2.88 \\ 
   \hline
p-val Friedman test & 0.000627 &  &  &  \\ 
  Adj. p-val Wilcoxon test & 0.00146 & 0.47 &  & 0.297 \\ 
   \hline
\hline
\end{tabular}
\endgroup
\caption{Results for simulated data with a sample size of n=500: Average test error rates for classification and test negative log-likelihoods for regression. 
    In parentheses are approximate standard deviations. 
    Below are average ranks of the methods over the different datasets (only considering datasets for which all four methods are run). 
    Further, a p-value of a Friedman test with an Iman and Davenport correction for comparing the different algorithms is reported.
    The last row shows Holm-Bonferroni corrected p-values of Wilcoxon signed-rank tests for pairwise 
    comparison of Newton boosting with the novel number of weighted samples parameter and the three alternative methods.} 
\label{results_simulated_smalldata}
\end{table}

\end{appendices}

\bibliographystyle{apa}
\end{document}